\documentclass[table]{ai2style/ai2}
\pdfoutput=1

\newif\ifallpoints
\allpointstrue

\usepackage{xspace}
\usepackage{pifont}
\usepackage{makecell}
\usepackage{adjustbox}
\usepackage{arydshln}
\usepackage{wrapfig}
\usepackage{enumitem}
\usepackage{amsmath}
\usepackage{amssymb}
\usepackage{fdsymbol}
\setlist[itemize]{leftmargin=*,itemsep=0em,parsep=0.3em,topsep=0.3em}

\newcommand{\benchmark}{\mbox{RoboTrack}}
\newcommand{\benchmarkreal}{\mbox{RoboTrack-Bench}}
\newcommand{\benchmarkdata}{\mbox{RoboTrack-15M}}
\newcommand{\modelname}{\mbox{RoboTracker}}
\newcommand{\modelnamewarp}{\mbox{RoboTracker-Warp}}
\newcommand{\proto}[2]{\ifallpoints#2\else#1\fi}

\usepackage{placeins} %
\usepackage{capt-of}

\definecolor{ai2pink}{HTML}{f0529c}
\definecolor{linkcolor}{RGB}{0, 0, 128}
\hypersetup{colorlinks=true, citecolor=linkcolor, linkcolor=linkcolor, urlcolor=linkcolor}

\newcommand{\huggingface}{\raisebox{-1.5pt}{\includegraphics[height=1.05em]{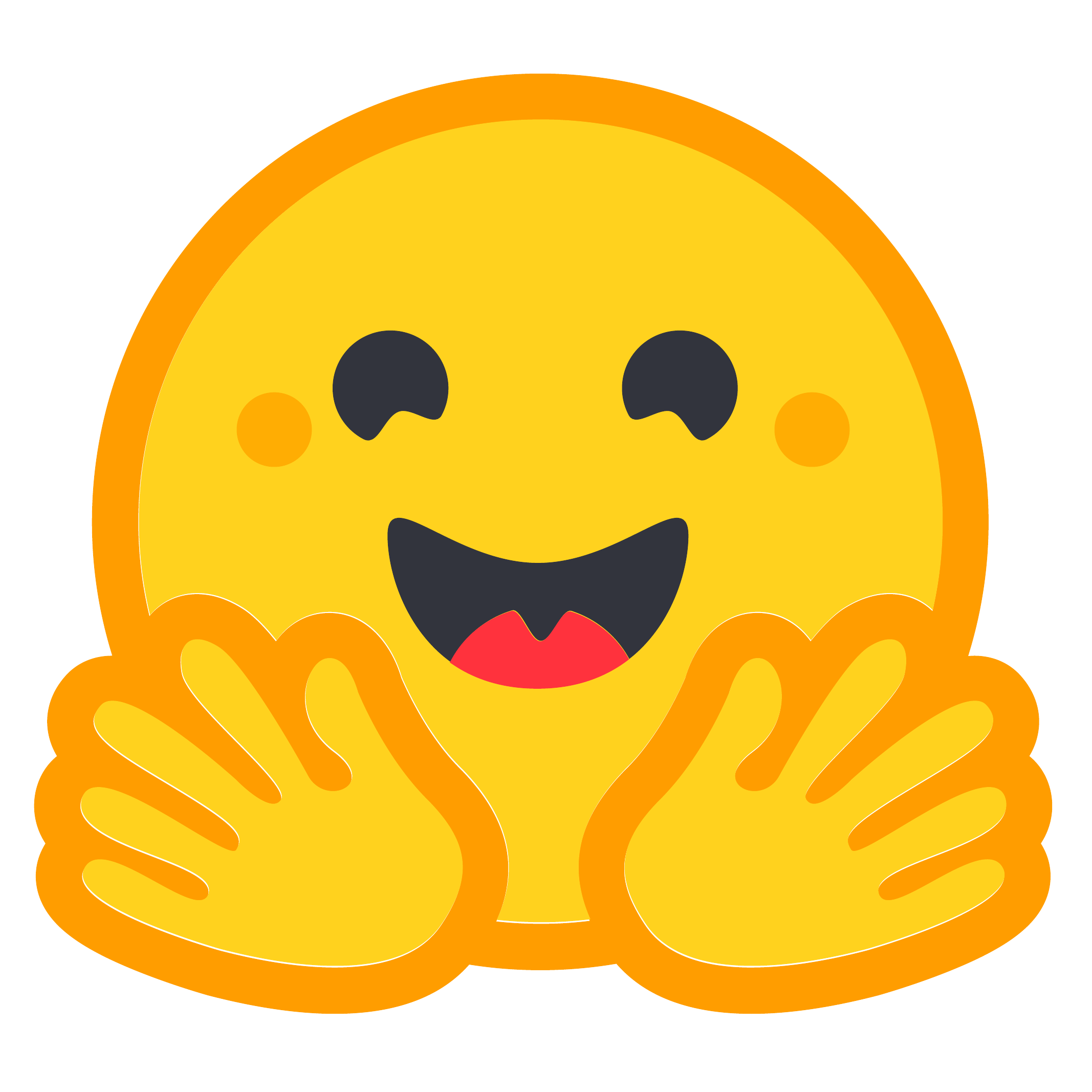}}\xspace}
\newcommand{\hfdataset}{\raisebox{-1.5pt}{\includegraphics[height=1.05em]{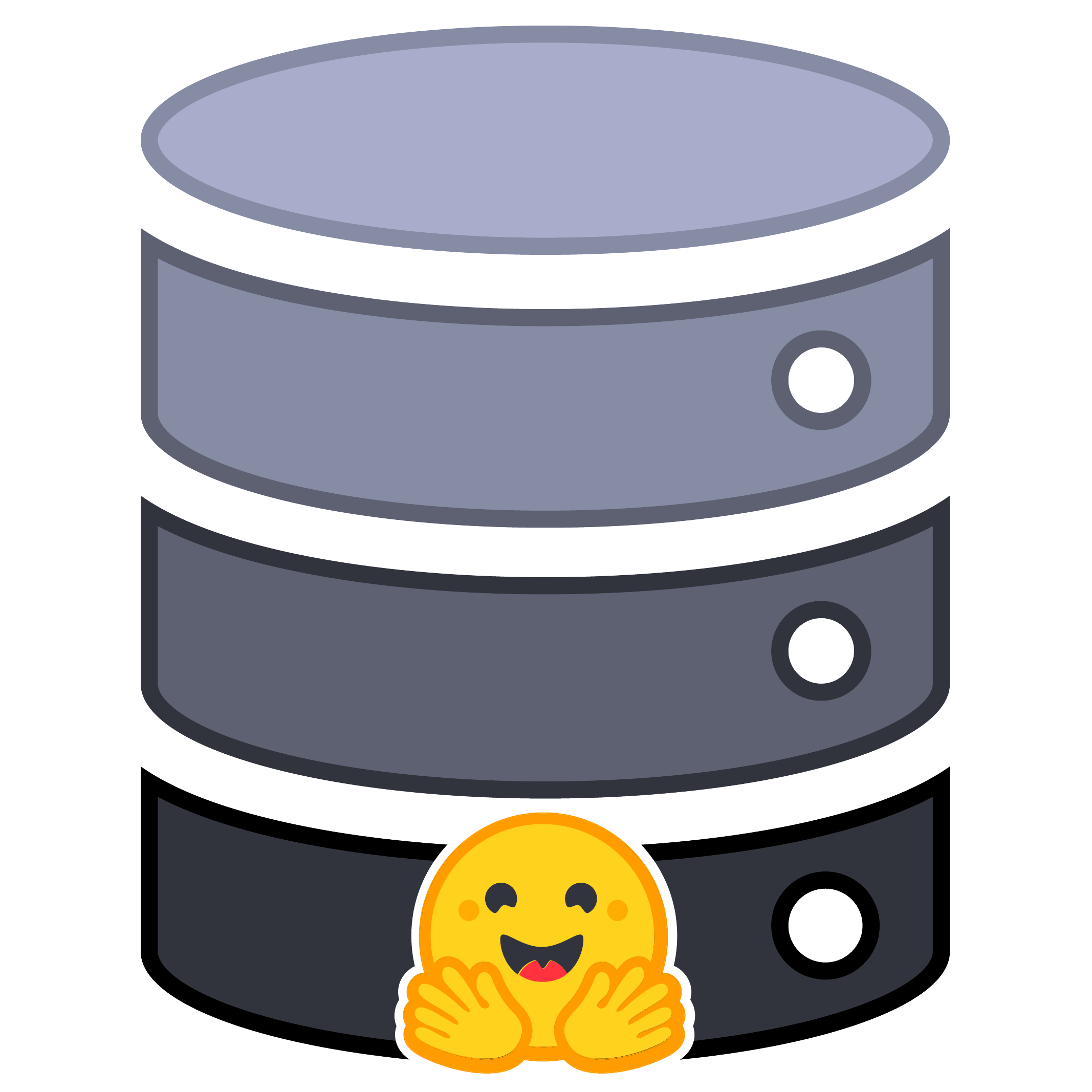}}\xspace}
\newcommand{\github}{\raisebox{-1.5pt}{\includegraphics[height=1.05em]{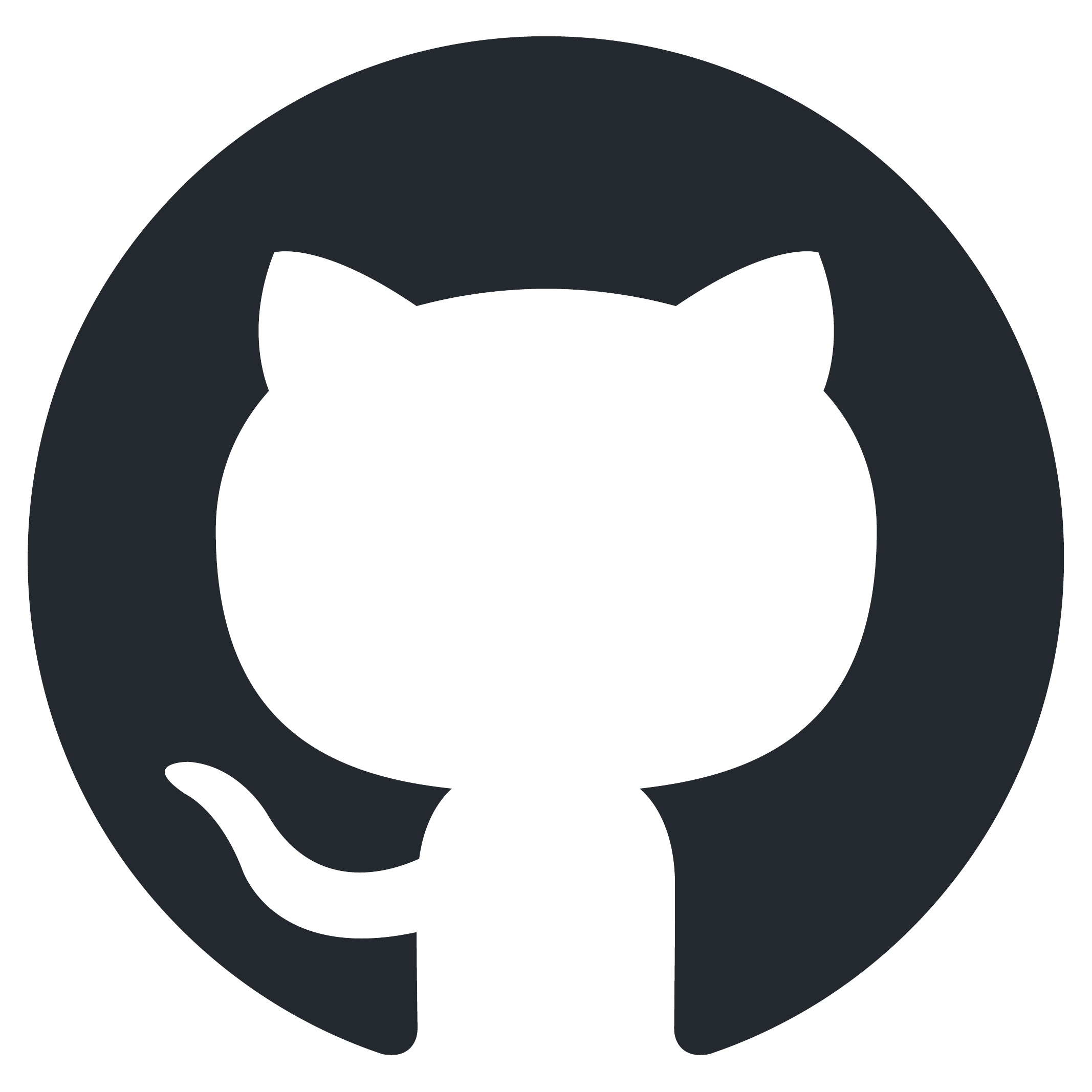}}\xspace}
\newcommand{\aitoo}{\raisebox{-1.5pt}{\includegraphics[height=1.05em]{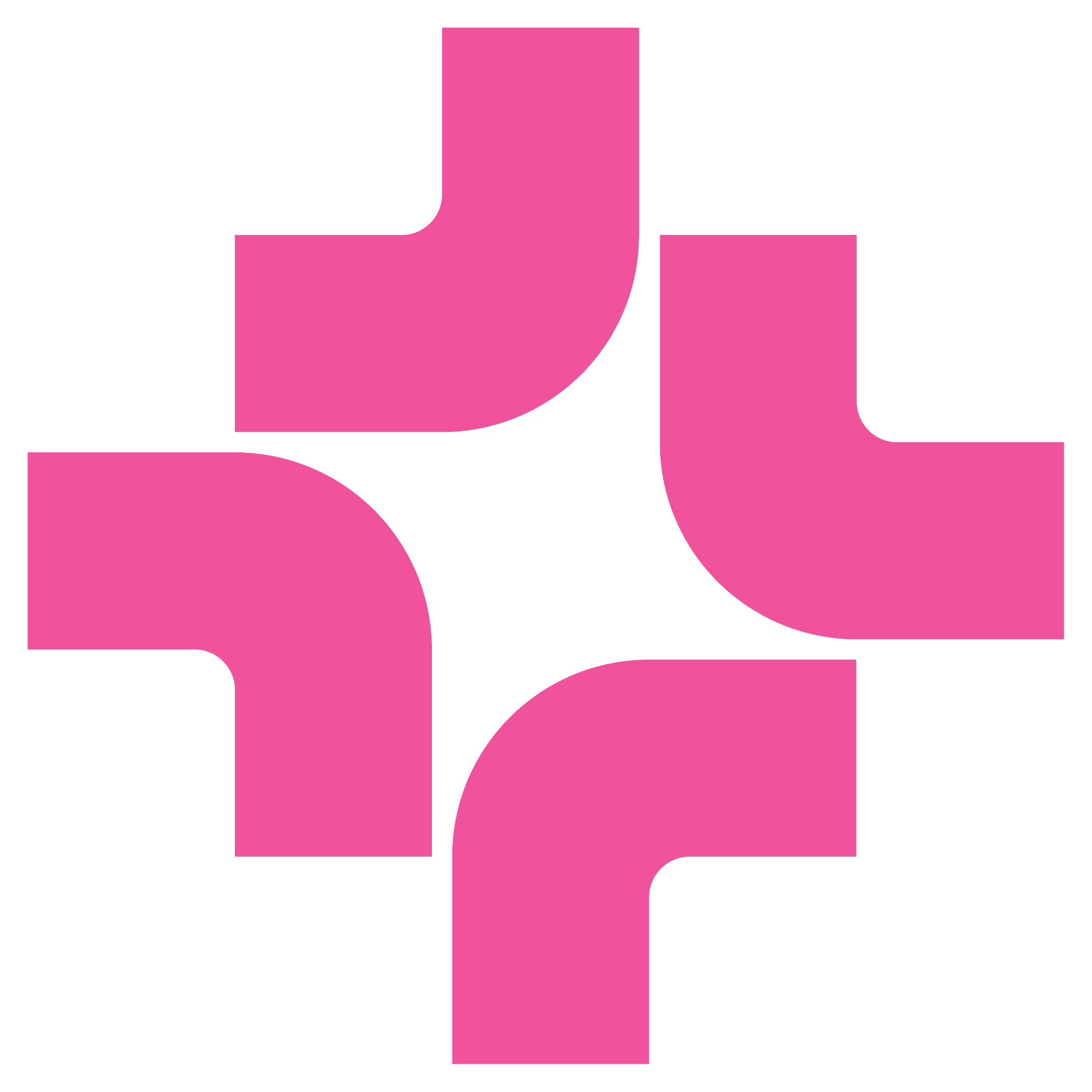}}\xspace}

\title{RoboTrack\\{\fontsize{18pt}{12pt}\selectfont Point Tracking for Real-World Robotic Manipulation}}

\authorOne[1,2]{Vincent Shao}
\authorOne[1]{Max Argus}
\authorOne[1]{Winson Han}
\authorOne[1]{Eli VanderBilt}
\authorOne[1]{Carissa Schoenick}
\authorTwo[2]{Ali Farhadi}
\authorTwo[2]{Ranjay Krishna}
\authorTwo[1,3]{Zhongzheng Ren}
\authorTwo[1]{Jae Sung Park}

\renewcommand\affiliationformat[2][]{{\normalsize #2$^{#1}$}}
\affiliation[1]{Allen Institute for AI}
\affiliation[2]{University of Washington}
\affiliation[3]{University of North Carolina at Chapel Hill}

\abstract{Point trackers turn video into motion trajectories that robotic systems can rely on to learn from demonstrations and guide their manipulation. However, we find that off-the-shelf trackers with strong performance on established benchmarks struggle with the varied interactions and viewpoints of real-world manipulation. To address this gap, we introduce \textbf{\benchmark{}}, a benchmark and training suite for point tracking in robotic manipulation. Our benchmark, \textbf{\benchmarkreal{}}, contains 307 real-world clips with 2,522 human-annotated tracks covering four robot embodiments and human manipulation from exocentric and egocentric views. Evaluation shows that many existing trackers achieve lower scores on \benchmarkreal{} than on prior robotics benchmarks, with even the strongest baseline achieving only 63.9 Average Jaccard (AJ), compared with 85.5 on RGB-Stacking and \proto{73.5}{70.5} on RoboTAP. To improve tracking in robotic manipulation, we extend the MolmoSpaces simulator to generate \textbf{\benchmarkdata{}}, a synthetic training set of manipulation-aligned point tracks. We find that fine-tuning on this data improves \benchmarkreal{} by up to 2.1 AJ without hurting general-video tracking and yields state-of-the-art results on all robotics benchmarks, reaching 66.0 AJ on \benchmarkreal{}. Lastly, we demonstrate how point tracking can be applied to a realistic robotics task in a wet lab setting. Using point tracks from our fine-tuned tracker, a robot can transfer liquid between two vials in 17 of 20 trials despite occlusions, compared with at most 8 for flow- and mask-based alternatives.
}

\metadata[\quad\huggingface Models:]{
\href{https://huggingface.co/allenai/RoboTracker}{\texttt{RoboTracker}} \quad
\href{https://huggingface.co/allenai/RoboTracker-Warp}{\texttt{RoboTracker-Warp}}}

\metadata[\vspace{.5em}\quad\hfdataset Data:]{
\href{https://huggingface.co/datasets/allenai/RoboTrack-Bench}{\texttt{RoboTrack-Bench}} \quad
\href{https://huggingface.co/datasets/allenai/RoboTrack-15M}{\texttt{RoboTrack-15M}}}

\metadata[\vspace{.5em}\quad\github Code:]{
\href{https://github.com/allenai/robotrack}{\texttt{https://github.com/allenai/robotrack}}}

\metadata[\vspace{.5em}\quad\aitoo Websites:]{
\href{https://robotrack.allen.ai}{\texttt{robotrack.allen.ai}} \quad
\href{https://allenai.org/blog/robotrack}{\texttt{Blog}}}

\begin{document}

\maketitle

\begin{figure}[t]
    \centering
    \includegraphics[width=\textwidth]{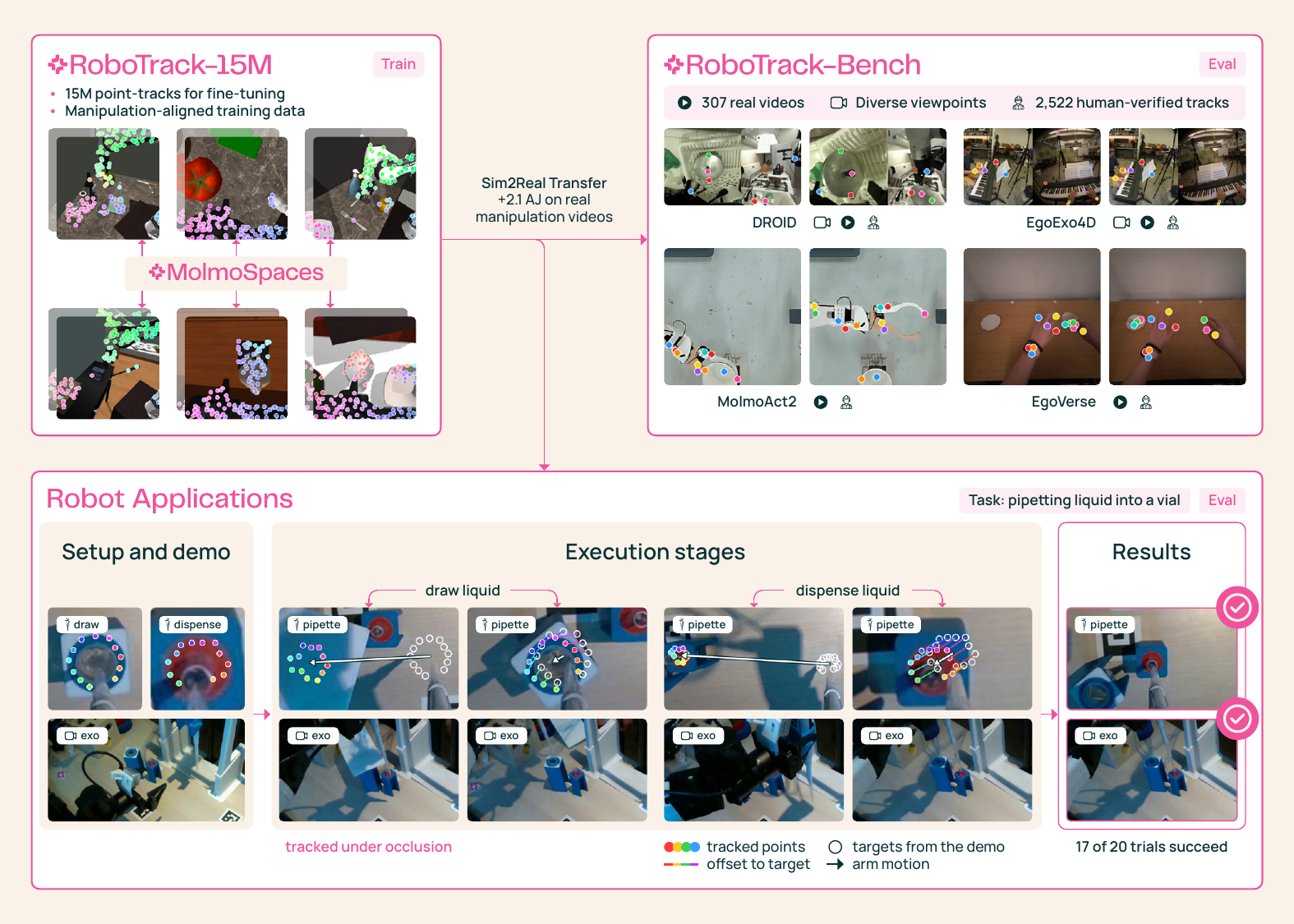}
    \caption{\textbf{\benchmark{}} is a data suite for point tracking in robotic manipulation. \textbf{\benchmarkdata{}} (top left) provides manipulation-aligned training data from MolmoSpaces that improves tracking on real robot videos. \textbf{\benchmarkreal{}} (top right) provides human-annotated point tracks on diverse in-the-wild manipulation videos. Lastly, we apply tracking to a real wet lab automation robotics task, where points from a single demo are tracked in the live camera and the arm is servoed until the points align with their targets in the demo.}
    \label{fig:teaser}
\end{figure}

\section{Introduction}

Point tracking estimates the motion of arbitrary image points through a video, across changes in viewpoint, deformation, and occlusion~\citep{harley2022particle,tapvid}. Modern point trackers are trained with synthetic data, where ground-truth trajectory annotations can be generated automatically from simulated scenes~\citep{greff2022kubric,zheng2023pointodyssey}. Despite relying heavily on synthetic supervision, these models generalize well to real-world videos~\citep{cho2024local,cotracker,cotracker3,bootstap}. Point tracking is particularly attractive for robotic manipulation, where tracks of the motion of physical surfaces provide a compact representation for grasping, pushing, placing, and other object interactions~\citep{bharadhwaj2024track2act,wen2023atm,xu2024im2flow2act,vecerik2024robotap,huang2024rekep,zhang2026molmomotion,zhang2026grounded}.

\begin{figure}[t]
    \centering
    \includegraphics[width=0.9\linewidth]{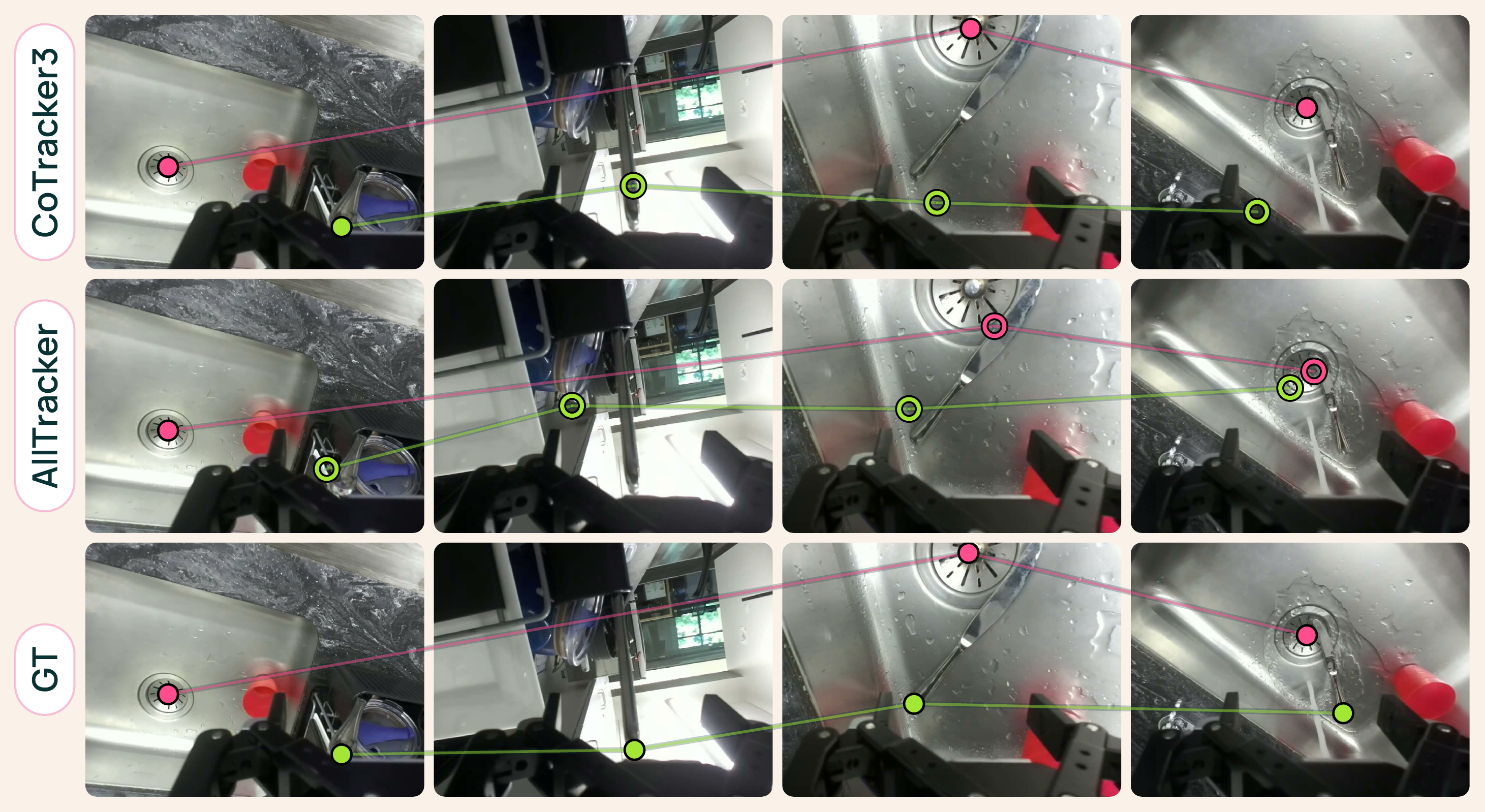}
    \caption{\textbf{Existing point tracking models in real manipulation videos.} Hollow points are predicted occluded. The state-of-the-art trackers lose correspondence with manipulated objects, such as the tip of a knife, following brief occlusions by the gripper.}
    \label{fig:motivation}
\end{figure}

However, manipulation in particular presents difficult tracking conditions: robot grippers occlude objects during interaction, objects move and rotate with the robot, and previously hidden points must be recovered when they reappear. Despite strong results on existing robotic tracking benchmarks, including RGB-Stacking~\citep{tapvid} and RoboTAP~\citep{vecerik2024robotap}, we find that current trackers frequently fail under these conditions (\Cref{fig:motivation}).

To evaluate this gap, we introduce \textbf{\benchmarkreal{}}, a benchmark containing 307 real manipulation videos and 2,522 human-verified point trajectories from DROID~\citep{khazatsky2024droid}, Ego-Exo4D~\citep{grauman2024egoexo4d}, EgoVerse~\citep{egoverse}, and MolmoAct2~\citep{molmoact2}, spanning diverse robots, objects, environments, and ego- and exocentric viewpoints. On \benchmarkreal{}, we find that \proto{every evaluated tracker performs worse  than on RoboTAP~\citep{vecerik2024robotap}, with drops of 6.7--10.1 Average Jaccard (AJ) points.}{every leading tracker performs worse than on RoboTAP~\citep{vecerik2024robotap}, with drops of 3.0--6.6 Average Jaccard (AJ) points.} The gaps between trackers also narrow, as CoWTracker~\citep{lai2026cowtracker}, the strongest baseline, leads CoTracker3 (online)~\citep{cotracker3} by \proto{4.0}{4.7} AJ on RoboTAP but by only 2.1 AJ on \benchmarkreal{}. Existing benchmarks thus overstate both how ready current trackers are for in-the-wild manipulation and how much they differ from one another.

We next investigate whether supervision with synthetic manipulation videos can narrow this gap. Existing synthetic datasets capture general object motion, rigid-body dynamics, and articulated characters~\citep{harley2022particle,greff2022kubric,zheng2023pointodyssey}, but do not specifically target robot-object interactions. We extend MolmoSpaces~\citep{kim2026molmospaces} to annotate simulated robotic manipulation with ground-truth point trajectories, producing \textbf{\benchmarkdata{}}. 

Our results show that \benchmarkdata{} transfers beyond the simulated environment. Starting from a Kubric-pretrained AllTracker, we replace its second training stage with just 10k steps on a mixture of Kubric and \benchmarkdata{}. Compared with training on Kubric alone, this improves tracking on all five evaluated benchmarks, including gains of 1.5 AJ on \benchmarkreal{}, 3.0 AJ on RGB-Stacking, 1.2 AJ on DAVIS, \proto{and 0.6 AJ each on RoboTAP and Kinetics}{0.7 AJ on RoboTAP, and 0.6 AJ on Kinetics}. The fine tuned model also matches or outperforms the released AllTracker, whose second stage runs for 400K steps, on all benchmarks except DAVIS. We also train using our data starting from the final AllTracker to produce a stronger model, which we call \textbf{\modelname{}}. The gains carry over to a different architecture as well; CoWTracker fine-tuned in the same way, which we call \textbf{\modelname{}-Warp}, improves \benchmarkreal{} by 2.1 AJ to 66.0 and achieves \textbf{state-of-the-art results on all robotics benchmarks}.

Lastly, we show an example application of point tracking to manipulation with a real robot in a wet lab setting in which a robot has to pipette liquid into a small vial, similar to RoboTAP \citep{vecerik2024robotap}. We find point tracks from the tracker fine-tuned on \benchmarkdata{} reliably guide the robot’s policy, even when the view of the task is partly occluded. Together, \benchmarkreal{} and \benchmarkdata{} provide an evaluation and training suite for studying how synthetic point trajectories transfer to real-world manipulation.

\section{Related Work}
\label{sec:related}

\textbf{Track any point.}
Visual tracking has been studied at several levels of granularity, from bounding boxes in driving, pedestrian, and surgical video~\citep{Wang2024MCTrackAU,Yamane2025MVTrajecterMP,nwoye2023cholectrack20}, to masks for video segmentation~\citep{ravi2024sam2,carion2025sam3segmentconcepts}, keypoints for human and animal pose~\citep{Sridhar2016RealTimeJT,deeplabcut,Pereira2022sleap}, and individual points~\citep{st4rtrack2025,Alumootil2025DePT3RJD,tapir,tapvid,cotracker,cotracker3,harley2025alltracker}.
The point formulation, introduced as Particle Video~\citep{Sand2006ParticleVL}, has received renewed attention through datasets and models for tracking arbitrary queried points~\citep{harley2022particle,tapvid}.
While closely related to optical flow~\citep{horn1981determining,dosovitskiy2015flownet,teed2020raft}, point-tracking additionally requires long-range association and visibility prediction.
Recent trackers build on architectures developed for flow and correspondence~\citep{teed2020raft,wang2024sea}.
Several use cost volumes with iterative refinement~\citep{harley2022particle,tapir,tapvid,cho2024local}, CoTracker~\citep{cotracker,cotracker3} tracks many points jointly, TAPNext~\citep{zholus2025tapnext} recasts tracking as next-token prediction, AllTracker~\citep{harley2025alltracker} estimates dense long-range correspondence for every pixel, and TAPIP3D~\citep{tapip3d} tracks points in 3D.

\textbf{Training data for point tracking.}
Because point tracks are hard to obtain on real video, trackers mostly train on synthetic data, such as FlyingThings++~\citep{harley2022particle,mayer2016large}, Kubric~\citep{greff2022kubric}, which renders objects falling under camera motion, and PointOdyssey~\citep{zheng2023pointodyssey}, which adds long human and animal motion. Despite this, recent trackers transfer well to real video, and further gains come from pseudo-labeling real footage~\citep{bootstap,cotracker3,realworldpseudo2026}, dense flow supervision~\citep{harley2025alltracker}, or stronger augmentation~\citep{jung2026tapnext++}. Tracks on the robot can be derived from joint states and calibrated cameras~\citep{tapvidmv}, but this covers only the robot and can mislabel occlusions caused by other objects.

\begin{table}[t]
\centering
\caption{\textbf{Point tracking benchmarks for robotic manipulation.} \benchmarkreal{} covers the most videos, embodiments, and occlusions on real manipulation with the least static points (move less than one pixel at 256-pixel height).}
\label{tab:benchmark_comparison}
\setlength{\tabcolsep}{4pt}
\scalebox{1.05}{%
\begin{tabular}{lccc}
\toprule
 & \textbf{RGB-Stacking} & \textbf{RoboTAP} & \textbf{\benchmarkreal{}} \\
\midrule
Source              & Synthetic & Real & Real \\
Videos              & 50  & 265 & 307 \\
Frames / video      & 250 & 272 & 223 \\
Points / video      & 30  & 43.7 & 8.2 \\
Resolution          & $256^2$ & $296{\times}222$ & $\geq\!640{\times}360$ \\
Camera views        & exo & exo, ego & exo, ego \\
Embodiments         & 1 & 1 & \textbf{4 + human} \\
Static points       & 37\% & 20\% & \textbf{2.2\%} \\
Occluded at least once & 48\% & 56\% & \textbf{57\%} \\
\bottomrule
\end{tabular}%
}
\end{table}

\textbf{Benchmarks for point tracking.}
TAP-Vid~\citep{tapvid} established the standard evaluation protocol on real videos from DAVIS~\citep{pont20172017} and Kinetics~\citep{kay2017kinetics}. PointOdyssey~\citep{zheng2023pointodyssey} broadens synthetic evaluation to diverse domains for long-term point tracking, while SynthVerse~\citep{zhao2026synthverse} includes embodied manipulation. Domain-specific benchmarks cover animals~\citep{biggs2018creatures,mathis2021pretraining}, surveillance~\citep{sundararaman2021tracking}, driving~\citep{balasingam2024drivetrack}, and egocentric video~\citep{darkhalil2025egopoints}, while ITTO~\citep{demler2025itto} emphasizes complex motion and re-identification after occlusion. TAPVid-3D~\citep{koppula2024tapvid} and WorldTrack~\citep{st4rtrack2025} extend evaluation to 3D trajectories. For robotics, TAP-Vid~\citep{tapvid} includes synthetic annotations on the RGB-Stacking dataset~\citep{lee2021beyond}, while RoboTAP~\citep{vecerik2024robotap} provides human-annotated tracks for real tabletop manipulation videos. These latter two benchmarks emphasize relatively controlled tabletop settings, with limited scene clutter, camera variation, and object diversity. Recent work such as TAPVid-MV~\citep{tapvidmv} evaluates multi-view tracking on DROID~\citep{khazatsky2024droid}, but the trajectories are derived only from the robot and static background. \Cref{tab:benchmark_comparison} contrasts existing robotics benchmarks with \benchmarkreal{}, highlighting differences in camera viewpoints, manipulation embodiments, and trajectory characteristics.

\textbf{Point tracking in robotics.}
Point tracks provide a useful representation of object motion for learning manipulation from demonstration videos~\citep{wen2023atm,bharadhwaj2024gen2act,haldar2025point,xu2024im2flow2act} and visual feedback for guiding execution. Track2Act~\citep{bharadhwaj2024track2act} uses point tracks to predict object trajectories, while RoboTAP~\citep{vecerik2024robotap} uses keypoint re-detection to transfer behavior from demonstrations to live execution. These uses extend to 3D, where tracks serve as manipulation plans~\citep{hung2026tripointr}, targets for dexterous control~\citep{kuang2026dex4d}, and supervision for dynamics prediction~\citep{huang2026pointworld}. Such applications depend on off-the-shelf trackers like CoTracker~\citep{cotracker3} to obtain the point tracks. \benchmarkreal{} exposes where these trackers fail on real manipulation videos, while we introduce \benchmarkdata{} as training data to address these failures.

\section{\benchmark{}: Benchmark and Training Data for Robotic Manipulation}
\label{sec:benchmark}

In this section, we detail the benchmark statistics and dataset creation framework for \benchmark{}.

\subsection{\benchmarkreal{}: Real Manipulation Videos}
\label{sec:real}

\begin{figure}[!ht]
  \centering
  \includegraphics[width=\textwidth]{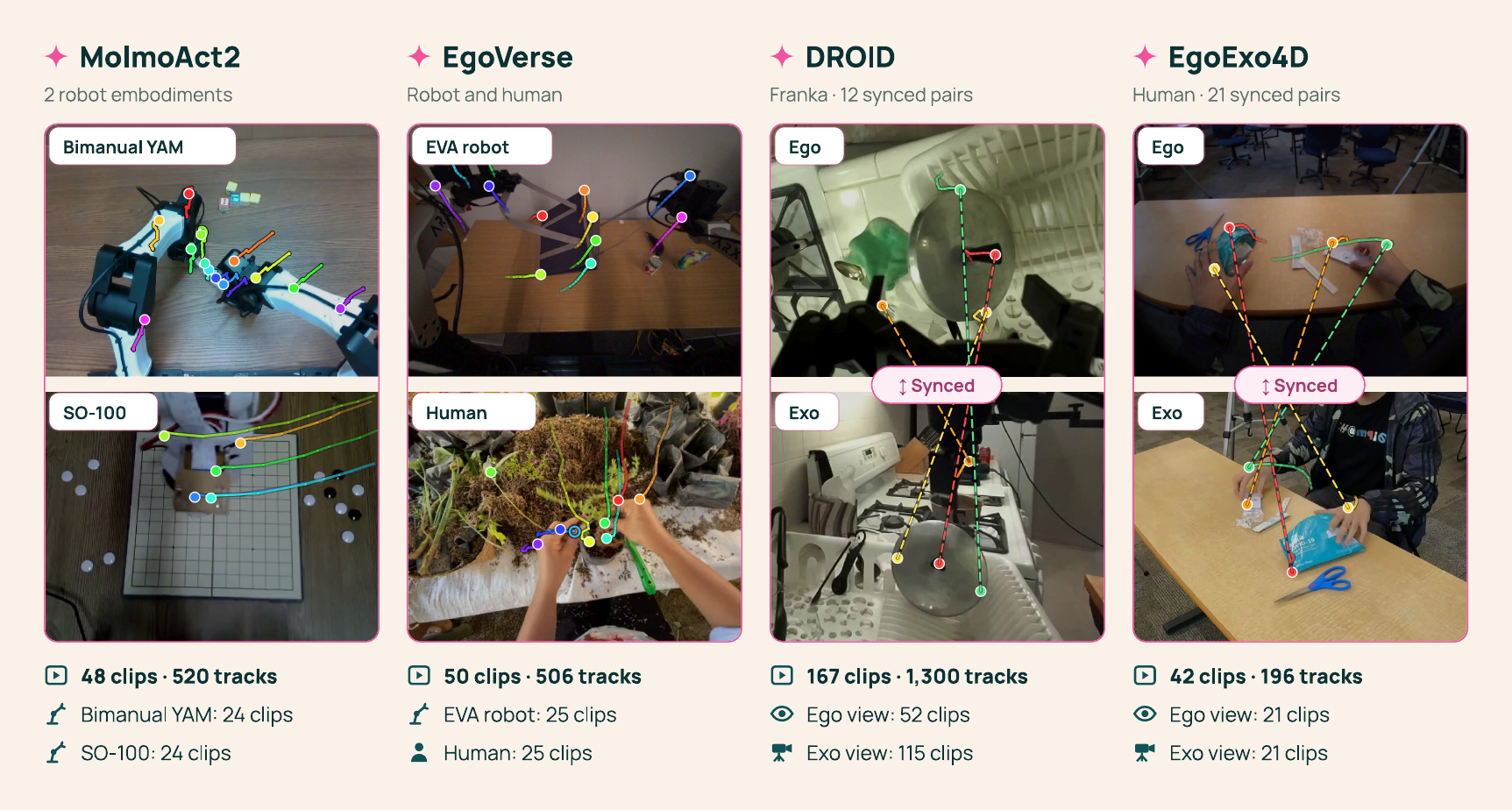}
  \caption{\textbf{\benchmarkreal{} at a glance.}
  Examples from each source dataset (307 clips and 2,522 human-verified tracks in total).
  MolmoAct2 and EgoVerse show one embodiment per panel. DROID and Ego-Exo4D show synchronized ego (top) and exo (bottom) views.}
  \label{fig:benchmark_examples}
\end{figure}

\textbf{Video sources} 
\benchmarkreal{} contains 307 human-annotated clips with 2,522 point tracks, drawn from DROID~\citep{khazatsky2024droid}, MolmoAct2~\citep{molmoact2}, EgoVerse~\citep{egoverse}, and Ego-Exo4D~\citep{grauman2024egoexo4d} (\Cref{fig:benchmark_examples}). The clips span four robot embodiments and human manipulation, selected for diversity in scenes, actions, and camera viewpoints. For subsets of DROID and Ego-Exo4D, we include synchronized egocentric and exocentric recordings of the same interaction, yielding 33 video pairs that support studying point correspondence across views. Resolutions and per-source statistics are in \Cref{tab:real_composition}.

\FloatBarrier %
\begin{table}[!tp]
\centering
\caption{\textbf{Composition of \benchmarkreal{}}. Exo and Ego count clips by camera view; Ego includes wrist-mounted cameras. 24 clips from DROID (synchronized pairs) and 42 clips from Ego-Exo4D form 33 synchronized exo–ego pairs.}
\label{tab:real_composition}
\small
\begin{tabular}{l l l r r r r}
\toprule
\textbf{Source} & \textbf{Embodiment} & \textbf{Resolution} & \textbf{Exo} & \textbf{Ego} & \textbf{Clips} & \textbf{Tracks} \\
\midrule
DROID (single camera)      & Franka       & $1280\times720$                    & 103 & 40  & 143 & 1,236 \\
DROID (synchronized pairs) & Franka       & $1280\times720$, $672\times376$    & 12  & 12  & 24  & 64 \\
MolmoAct2                 & bimanual YAM & $640\times360$                     & 23  & 1   & 24  & 263 \\
MolmoAct2                 & SO-100       & $640\times480$                     & 14  & 10  & 24  & 257 \\
EgoVerse                  & EVA          & $640\times480$                     & 25  & 0   & 25  & 265 \\
EgoVerse                  & Human        & $640\times360$ to $640\times480$   & 0   & 25  & 25  & 241 \\
Ego-Exo4D                  & Human        & $1920\times1080$, $1408\times1408$ & 21  & 21  & 42  & 196 \\
\midrule
\textbf{Total}            &              &                                    & \textbf{198} & \textbf{109} & \textbf{307} & \textbf{2,522} \\
\bottomrule
\end{tabular}
\end{table}

\textbf{Human annotation pipeline.}
We follow TAP-Vid~\citep{tapvid} in the design of our human annotation pipeline. Annotators are instructed to select any moving point in the video, with no restriction on object categories, and are encouraged to select points that have interesting motion or occlusions. 

\textit{Step 1: Point selection.} Annotators first watch the entire video and choose five points to annotate. Since exocentric robotics videos tend to contain many static points, we instruct annotators to only choose points which move at some point in the video. This eliminates background points and static objects. Each point is selected on the first frame in which it appears, which becomes the query point and frame in \benchmarkreal{}. After selecting a point, annotators label it in one of three categories based on the object that the point lies on: Embodiment (Robot or Human), Manipulated Object, Static Object (Camera Motion). 

\textit{Step 2: Track refinement.} Similar to TAP-Vid~\citep{tapvid}, we provide a track assist algorithm that uses SEA-RAFT~\citep{wang2024sea} to convert sparsely chosen points into dense tracks that follow the motion of the point. For each of the five points, annotators review the initial track produced by the tracking assistant and add refinement points to fix its drift, making further corrections as needed. Annotators also mark segments in which the point is occluded; they are not required to track the point during occluded sections. See \Cref{appendix:track_assist} for further details on the tracking assistant.

\textit{Step 3: Verification.} Finally, in a separate verification pass, annotators check every frame for localization and visibility errors, making minor corrections where needed and accepting only fully correct tracks.

\begin{figure}[t]
    \centering
    \includegraphics[width=\linewidth]{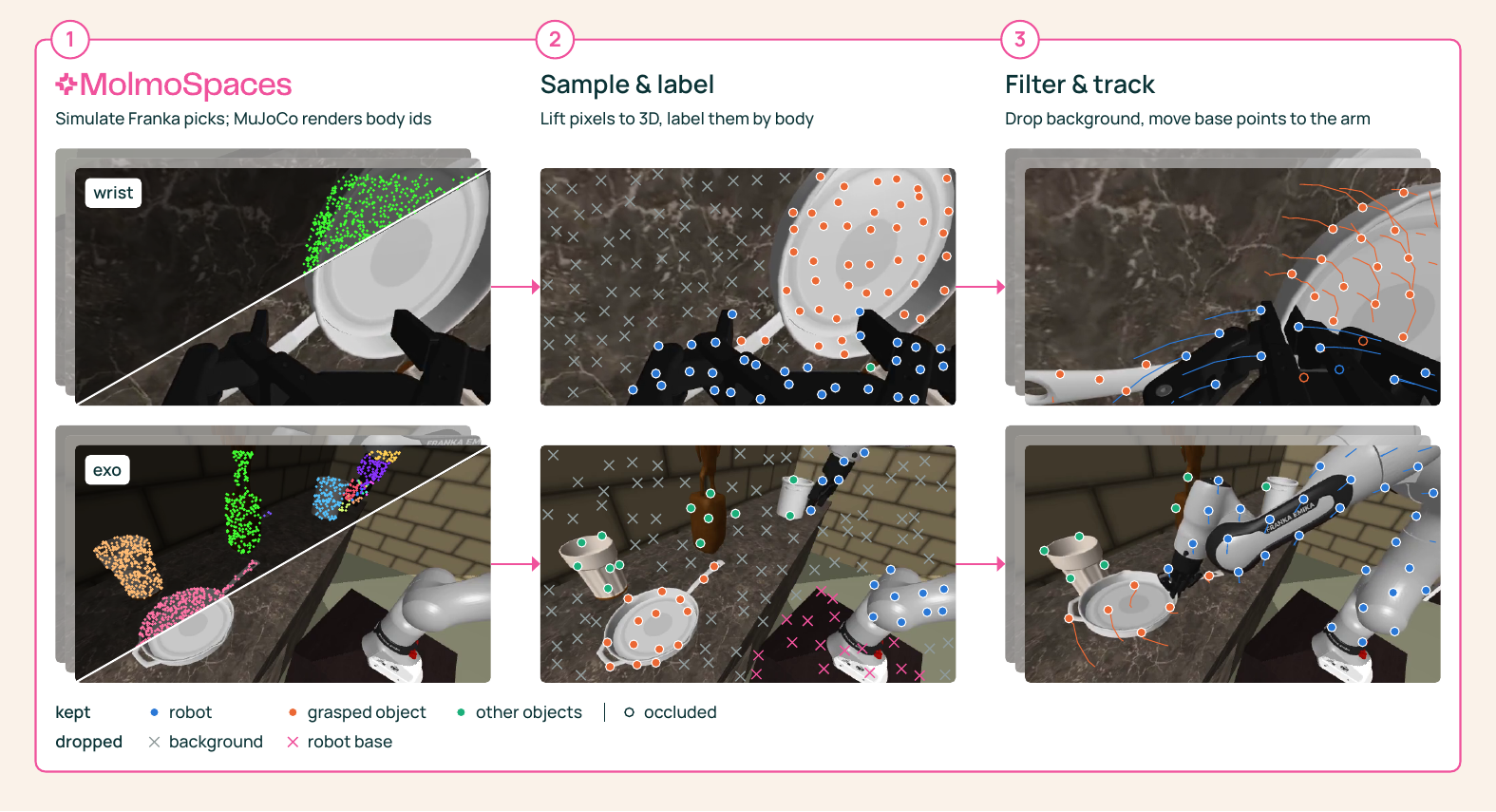}
    \caption{\textbf{Data generation pipeline of \benchmarkdata{}.} Objects, indoor scenes, and robot grasps are obtained from MolmoSpaces and rendered in MuJoCo. Point tracks are obtained by unprojecting sampled pixels to 3D and projecting them into every frame via camera intrinsics. Background and robot-base points are filtered out, and visibility is determined by a depth test.}
    \label{fig:datagen}
\end{figure}

\subsection{\benchmarkdata{}: Synthetic Simulation Data}
\label{sec:sim}

MolmoSpaces~\citep{kim2026molmospaces} is a robotics simulation ecosystem for generating training and testing data with diverse scenes, objects, and actions. We augment the MolmoSpaces data generation pipeline to produce 2D point tracks for the Tracking Any Point task (\Cref{fig:datagen}). 

\textbf{Scene generation.}
We use the MolmoSpaces scene generation as is, making no changes to how the robot picks up objects or to which indoor environments and objects are chosen. The houses are selected from ProcTHOR~\citep{deitke2022}, each with differing layouts of walls, floors, furniture, and lighting. The robot, a Franka arm, is placed in a valid location, and objects are spawned near the robot for manipulation. We use MuJoCo Classic for rendering and physics simulation. Two cameras capture different perspectives: a wrist camera on the gripper, and an exocentric camera that sees the robot arm and the objects being manipulated.

\textbf{Obtaining point tracks.}
One straightforward way to obtain point tracks from simulation is to use mesh vertices, following PointOdyssey~\citep{zheng2023pointodyssey}. However, we found that this produces ambiguous points near object edges, which appear to be off the object but are meant to be on it. Additionally, many of our objects are low-poly, so vertex sampling severely limits where points can be placed. Instead, we use an approach closer to Kubric~\citep{greff2022kubric}: we sample points from visible object pixels, unproject them to 3D using rendered depth, and store their positions in each body's local coordinate frame. This allows us to sample from every object pixel in the video, vastly increasing the diversity of our points.

\textbf{\benchmarkdata{}.}
The released training set covers 300 ProcTHOR houses with 5 pick episodes each, recorded from two cameras, giving 3,000 videos with 5,000 tracks each (15M tracks in total). Candidate pixels are drawn every 20 frames, uniformly over the visible pixels of movable objects and robot links, and none on walls, floors, or fixed furniture. See \Cref{tab:data_stats} in the Appendix for dataset statistics.

\begin{figure*}[t]
    \centering
    \includegraphics[width=0.9\textwidth]{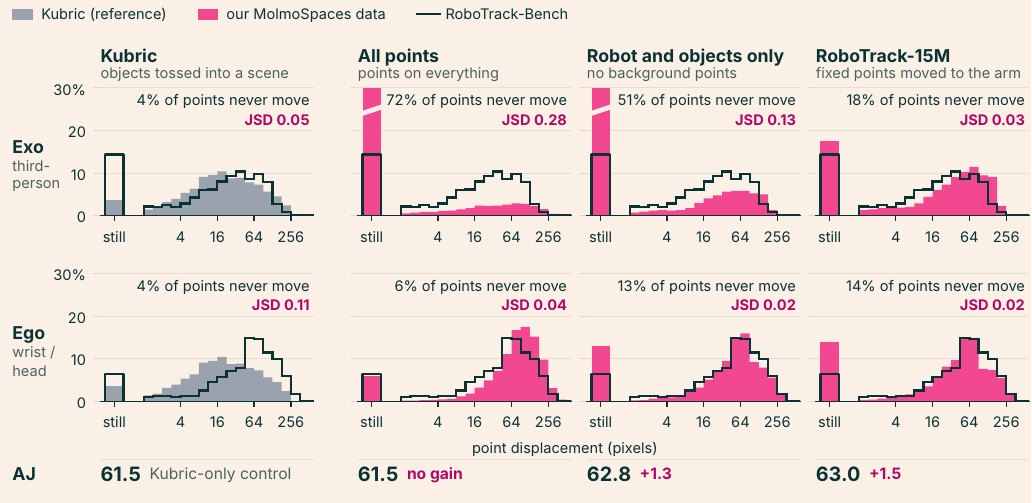}
    
    \caption{\textbf{Effect of point selection on point motion in our training data, compared with \benchmarkreal{}.} Each histogram shows how far points move on exo (top) and ego (bottom) cameras. The x-axis is the largest distance each point moves from its first visible position throughout the clip, in pixels at 256-pixel image height, and the y-axis is the percentage of points. Similarity to \benchmarkreal{} is measured with the Jensen-Shannon divergence (JSD). While Kubric closely matches exo videos (JSD 0.05), it shows low resemblance with ego videos (JSD 0.11). AJ is measured on \benchmarkreal{} after fine-tuning AllTracker on each dataset mixed with Kubric (\Cref{sec:data_ablation}).}
    \label{fig:data_histograms}
\end{figure*}

\paragraph{Selecting supervised points.}
We study how the selection of supervised points affects sim-to-real transfer by varying the proportion of tracks on robot surfaces, manipulated objects, and backgrounds. Our first attempt sampled points across the scene, but fine-tuning AllTracker on a mix of Kubric and this data (\Cref{sec:data_ablation}) brought no improvement over fine-tuning on Kubric alone. To diagnose this, we compare the point motion distribution of our data with that of \benchmarkreal{}. For each point, we measure its maximum displacement from its first visible position over 56 frames at 15 fps, normalized to an image height of 256 pixels, and compute the Jensen--Shannon divergence (JSD) between the resulting distributions in Figure~\ref{fig:data_histograms}. Our first attempt shows a large divergence from \benchmarkreal{} on exo videos (JSD 0.28), as most sampled points lie on static background. To close this sim-to-real gap, we first restrict points to the robot and manipulated objects, which aligns our ego videos with \benchmarkreal{}. We then move points on the fixed robot base onto the moving arm, which aligns the exo videos as well (JSD 0.03). The resulting \benchmarkdata{} achieves the highest AJ on our benchmark. Note that Kubric closely matches the exo videos (JSD 0.05) but diverges on the ego videos (JSD 0.11). Its objects move on their own in front of a static camera, whereas in robot manipulation the camera often moves with the robot, which may make Kubric less suited for manipulation tasks.

\section{Experiments}
\label{sec:experiments}
\ifallpoints\definecolor{oursbg}{HTML}{FAD9E8}

\begin{table*}[t]
\centering
\caption{\textbf{Point tracking results on robotic manipulation and general videos.} Higher is better for all metrics. Best results are in bold and second best are underlined, and our models are shaded. Among the off-the-shelf models, CoWTracker~\cite{lai2026cowtracker} performs best on average across the robotics benchmarks, followed by AllTracker~\cite{harley2025alltracker}. \modelname{} and \modelname{}-Warp are AllTracker and CoWTracker fine-tuned with \benchmarkdata{}, improving AJ over their base models on all benchmarks except RoboTAP for \modelname{} and DAVIS for \modelname{}-Warp. For training data, 15k and 15M are the real videos pseudo-labeled by CoTracker3~\cite{cotracker3} and BootsTAPNext~\cite{bootstap}, and mix is AllTracker's mix of point tracking and optical flow datasets in stage 2 training.}
\label{tab:results}
\setlength{\tabcolsep}{3.5pt}
\resizebox{\textwidth}{!}{
\begin{tabular}{l >{\footnotesize}l r ccc ccc ccc ccc >{\hspace{1.2em}}c cc ccc}
\toprule
&  &  & \multicolumn{12}{c}{{Robotic manipulation}} & \multicolumn{6}{c}{{General}} \\
\cmidrule(lr){4-15} \cmidrule(l{1.9em}r){16-21}
&  &  & \multicolumn{3}{c}{\textbf{\benchmarkreal{}}} & \multicolumn{3}{c}{\textbf{RGB-Stack}} & \multicolumn{3}{c}{\textbf{RoboTAP}} & \multicolumn{3}{c}{\textbf{Avg.}} & \multicolumn{3}{c}{\textbf{Kinetics}} & \multicolumn{3}{c}{\textbf{DAVIS}} \\
\cmidrule(lr){4-6} \cmidrule(lr){7-9} \cmidrule(lr){10-12} \cmidrule(lr){13-15} \cmidrule(l{1.9em}r){16-18} \cmidrule(lr){19-21}
Model & \multicolumn{1}{l}{Train. Data} & Params & AJ & $\delta_{\text{vis}}$ & OA & AJ & $\delta_{\text{vis}}$ & OA & AJ & $\delta_{\text{vis}}$ & OA & AJ & $\delta_{\text{vis}}$ & OA & AJ & $\delta_{\text{vis}}$ & OA & AJ & $\delta_{\text{vis}}$ & OA \\
\midrule
CoTracker3 (offline) & Kubric & 25M
& 59.8 & 73.2 & 84.6
& 76.6 & 86.4 & 90.8
& 59.2 & 73.7 & 85.9
& 65.2 & 77.8 & 87.1
& 53.4 & 66.4 & 86.1
& 63.6 & 76.7 & 88.2 \\

CoTracker3 (offline) & Kubric+15k & 25M
& 60.2 & 72.6 & 85.8
& 74.4 & 85.0 & 91.6
& 64.7 & 78.0 & 89.4
& 66.5 & 78.5 & 88.9
& 54.8 & 67.8 & 87.6
& 64.8 & 76.9 & \underline{91.7} \\

CoTracker3 (online) & Kubric+15k & 25M
& 61.8 & 74.9 & 87.3
& 70.7 & 83.3 & 91.4
& 65.8 & 78.9 & 90.6
& 66.1 & 79.0 & 89.8
& 55.1 & 68.1 & 88.3
& 64.4 & 76.8 & 90.6 \\

TAPNext & Kubric & 194M
& 55.1 & 69.7 & 87.2
& 75.0 & 85.4 & \textbf{95.2}
& 58.7 & 72.2 & 87.8
& 62.9 & 75.8 & 90.1
& 48.8 & 63.3 & 85.6
& 62.8 & 77.0 & 90.5 \\

BootsTAPNext & Kubric+15M & 194M
& 60.2 & 73.5 & 87.1
& 65.5 & 78.2 & 86.8
& 63.2 & 74.5 & 88.4
& 62.9 & 75.4 & 87.5
& 52.4 & 65.9 & 86.3
& \textbf{66.6} & \textbf{79.5} & \textbf{92.1} \\

AllTracker & Kubric & 17M
& 61.4 & 74.5 & 83.9
& 80.7 & 90.1 & 91.7
& 67.5 & 79.3 & 90.8
& 69.9 & 81.3 & 88.8
& 55.8 & 67.9 & 86.9
& 62.8 & 76.2 & 87.2 \\

AllTracker & Kubric+mix & 17M
& 62.6 & 75.7 & 86.9
& 81.4 & 90.1 & 92.9
& 68.5 & 80.8 & 91.7
& 70.8 & 82.2 & 90.5
& 56.3 & 69.0 & 88.8
& 64.3 & 77.1 & 89.5 \\

CoWTracker & Kubric & $\sim$1B
& \underline{63.9} & \underline{76.1} & \underline{88.5}
& \underline{85.5} & \underline{93.0} & \underline{94.8}
& \underline{70.5} & \underline{81.2} & \underline{93.9}
& \underline{73.3} & \underline{83.4} & \underline{92.4}
& \underline{57.6} & \underline{69.9} & \textbf{90.0}
& \underline{66.1} & 78.4 & 91.2 \\

\midrule

\rowcolor{oursbg}\modelname{} & Kubric+mix+\benchmarkdata{} & 17M
& 63.4 & \underline{76.1} & 86.2
& 83.2 & 90.9 & 93.5
& 68.4 & 80.6 & 91.3
& 71.7 & 82.5 & 90.3
& 56.6 & 69.0 & 87.6
& 64.6 & 77.4 & 89.1 \\

\rowcolor{oursbg}\modelname{}-Warp & Kubric+\benchmarkdata{} & $\sim$1B
& \textbf{66.0} & \textbf{77.9} & \textbf{89.0}
& \textbf{85.9} & \textbf{93.2} & 94.5
& \textbf{71.6} & \textbf{81.9} & \textbf{94.2}
& \textbf{74.5} & \textbf{84.4} & \textbf{92.6}
& \textbf{58.0} & \textbf{70.3} & \underline{89.7}
& 65.9 & \underline{78.9} & 89.9 \\

\bottomrule
\end{tabular}}
\end{table*}
\else

\begin{table*}[t]
\centering
\caption{\textbf{Point tracking results on robotic manipulation and general videos.} Higher is better for all metrics. Among the off-the-shelf models, CoWTracker performs best on average across the robotics benchmarks, followed by AllTracker. \modelname{} and \modelname{}-Warp are AllTracker and CoWTracker fine-tuned with \benchmarkdata{}, improving AJ over their base models on all benchmarks but one (DAVIS for \modelname{}-Warp.}
\label{tab:results}
\resizebox{\textwidth}{!}{
\begin{tabular}{l l r ccc ccc ccc ccc @{\hspace{1.5em}} ccc ccc}
\toprule
&  &  & \multicolumn{12}{c}{{Robotic manipulation}} & \multicolumn{6}{c}{{General}} \\
\cmidrule(lr){4-15} \cmidrule(lr){16-21}
&  &  & \multicolumn{3}{c}{\textbf{\benchmarkreal{}}} & \multicolumn{3}{c}{\textbf{RGB-Stack}} & \multicolumn{3}{c}{\textbf{RoboTAP}} & \multicolumn{3}{c}{\textbf{Avg.}} & \multicolumn{3}{c}{\textbf{Kinetics}} & \multicolumn{3}{c}{\textbf{DAVIS}} \\
\cmidrule(lr){4-6} \cmidrule(lr){7-9} \cmidrule(lr){10-12} \cmidrule(lr){13-15} \cmidrule(lr){16-18} \cmidrule(lr){19-21}
Model & Train. Data & Params & AJ & $\delta_{\text{vis}}$ & OA & AJ & $\delta_{\text{vis}}$ & OA & AJ & $\delta_{\text{vis}}$ & OA & AJ & $\delta_{\text{vis}}$ & OA & AJ & $\delta_{\text{vis}}$ & OA & AJ & $\delta_{\text{vis}}$ & OA \\
\midrule
CoTracker3 (offline) & Kubric & 25M
& 59.5 & 72.9 & 84.7
& 76.6 & 86.4 & 90.8
& 66.2 & 79.3 & 89.6
& 67.4 & 79.5 & 88.4
& 57.6 & 70.6 & 88.7
& 63.6 & 76.7 & 88.2 \\

CoTracker3 (offline) & Kubric+15k & 25M
& 60.2 & 72.5 & 85.9
& 74.4 & 85.0 & 91.6
& 70.3 & 81.8 & 92.2
& 68.3 & 79.7 & 89.9
& 59.0 & 71.7 & 89.7
& 64.8 & 76.9 & \underline{91.7} \\

CoTracker3 (online) & Kubric+15k & 25M
& 62.1 & 74.7 & 87.1
& 70.7 & 83.3 & 91.4
& 69.5 & 81.7 & 92.1
& 67.4 & 79.9 & 90.2
& 58.5 & 71.8 & 89.8
& 64.4 & 76.8 & 90.6 \\

TAPNext & Kubric & 194M
& 54.8 & 69.4 & 87.1
& 75.0 & 85.4 & \textbf{95.2}
& 63.3 & 77.2 & 89.3
& 64.4 & 77.3 & 90.5
& 52.9 & 67.9 & 87.4
& 62.8 & 77.0 & 90.5 \\

BootsTAPNext & Kubric+15M & 194M
& 59.9 & 73.3 & 87.1
& 65.5 & 78.2 & 86.8
& 67.6 & 78.9 & 90.2
& 64.3 & 76.8 & 88.1
& 56.0 & 69.7 & 88.4
& \textbf{66.6} & \textbf{79.5} & \textbf{92.1} \\

AllTracker & Kubric & 17M
& 61.1 & 74.2 & 83.9
& 80.7 & 90.1 & 91.7
& 71.1 & 82.3 & 92.4
& 71.0 & 82.2 & 89.3
& 59.4 & 71.3 & 89.3
& 62.8 & 76.2 & 87.2 \\

AllTracker & Kubric+mix & 17M
& 62.3 & 75.4 & 86.9
& 81.4 & 90.1 & 92.9
& 71.9 & 83.3 & 93.1
& 71.9 & 83.0 & 91.0
& 59.2 & 72.3 & 90.4
& 64.3 & 77.1 & 89.5 \\

CoWTracker & Kubric & $\sim$1B
& \underline{63.7} & \underline{75.9} & \underline{88.6}
& \underline{85.5} & \underline{93.0} & \underline{94.8}
& \underline{73.5} & \underline{83.7} & \underline{94.6}
& \underline{74.2} & \underline{84.2} & \underline{92.7}
& \underline{60.8} & \underline{73.1} & \textbf{91.5}
& \underline{66.1} & 78.4 & 91.2 \\

\midrule

\modelname{} & Kubric+mix+\benchmarkdata{} & 17M
& 63.2 & \underline{75.9} & 86.4
& 83.2 & 90.9 & 93.5
& 72.2 & 83.4 & 92.7
& 72.9 & 83.4 & 90.9
& 60.0 & 72.3 & 89.8
& 64.6 & 77.4 & 89.1 \\

\modelname{}-Warp & Kubric+\benchmarkdata{} & $\sim$1B
& \textbf{65.8} & \textbf{77.7} & \textbf{89.1}
& \textbf{85.9} & \textbf{93.2} & 94.5
& \textbf{74.3} & \textbf{84.3} & \textbf{94.8}
& \textbf{75.3} & \textbf{85.1} & \textbf{92.8}
& \textbf{61.1} & \textbf{73.4} & \underline{91.3}
& 65.9 & \underline{78.9} & 89.9 \\

\bottomrule
\end{tabular}}
\end{table*}\fi
We first evaluate baseline tracking models on our benchmark alongside existing ones. Then we ablate different training data compositions and compare performance on robotics benchmarks.

\textbf{Baseline models.}
We evaluate a range of state-of-the-art point trackers. For CoTracker3~\citep{cotracker3}, we include the Kubric-only model and a variant fine-tuned on 15k pseudo-labeled real videos.
TAPNext and BootsTAPNext~\citep{zholus2025tapnext} share an architecture but are trained on Kubric alone and additionally fine-tuned on 15M real videos, respectively. AllTracker~\citep{harley2025alltracker} is a dense tracker with similar architecture to CoTracker3, trained on Kubric and then on a mixture of optical flow and point tracking datasets. CoWTracker~\citep{lai2026cowtracker} replaces cost volumes with feature warping on top of a pretrained VGGT~\citep{wang2025vggt} backbone and is trained on Kubric.

\textbf{Evaluation protocol.}
We use the standard TAP-Vid~\citep{tapvid} metrics: Occlusion Accuracy (OA), $\delta_\text{vis}$, and Average Jaccard (AJ). All metrics are computed after rescaling predictions and ground truth to $256\times256$. We evaluate every tracker ourselves, each at its own input resolution: CoTracker3, TAPNext, and BootsTAPNext receive frames resized to $256\times256$, AllTracker receives $384\times512$ frames, and CoWTracker $336\times560$ frames. 

Following TAP-Vid~\citep{tapvid} and CoTracker3~\citep{cotracker3}, we evaluate in the ``first query'' mode, where query points are sampled from the first frame in which they become visible. \proto{On RoboTAP and Kinetics we follow AllTracker's protocol~\citep{harley2025alltracker} for every tracker: videos are trimmed to at most 600 frames and only the points visible at the start frame are tracked.}{On RoboTAP and Kinetics we evaluate every annotated point over the full length of every video, unlike AllTracker's protocol~\citep{harley2025alltracker}, which trims videos to at most 600 frames and tracks only the points visible at the start frame.} For each model, we follow the inference settings outlined in their papers.
\subsection{Benchmark Results}
\label{sec:benchmark_results}

\paragraph{Performance on existing robotics benchmarks does not transfer to in-the-wild manipulation.}
\Cref{tab:results} compares state-of-the-art point trackers on \benchmarkreal{} and four established benchmarks. Every evaluated model \proto{}{except CoTracker3 (offline, Kubric) }scores lower on \benchmarkreal{} than on RoboTAP, with the strongest baseline, CoWTracker, dropping from \proto{73.5}{70.5} on RoboTAP to 63.9 AJ on \benchmarkreal{}. This suggests that existing robotics benchmarks do not fully capture the challenges of diverse real-world manipulation that \benchmarkreal{} covers  with a broader range of objects, environments, and embodiments. Interestingly, AllTracker outperforms CoTracker3 (online) by \proto{2.4}{2.7} AJ on RoboTAP, but its lead shrinks to just 0.8 AJ on \benchmarkreal{}, and Kubric-only AllTracker falls behind it (61.4 vs.\ 61.8 AJ). Rankings can even reverse on robotics benchmarks, as TAPNext outperforms BootsTAPNext by 9.5 AJ on the synthetic RGB-Stacking but trails it by 5.1 AJ on \benchmarkreal{}. 

\paragraph{Strong general-video tracking performance does not guarantee robust manipulation tracking.} BootsTAPNext ranks first on DAVIS, yet falls significantly behind on \benchmarkreal{}, 3.7 AJ behind CoWTracker. It even falls behind the Kubric-only AllTracker on \benchmarkreal{} (60.2 vs.\ 61.4 AJ), despite outperforming it on DAVIS and being additionally trained on 15M pseudo-labeled real videos. This suggests that gains from large-scale real-video supervision alone do not necessarily carry over to manipulation, and we hypothesize that general web videos rarely contain the close-range hand–object interactions and occlusions common in manipulation. 

\paragraph{Fine-tuning on \benchmarkdata{} improves performance across benchmarks.}
To test whether simulated manipulation data can close this gap, we fine-tune AllTracker and CoWTracker on a mix of Kubric and \benchmarkdata{}, which we call \modelname{} and \modelname{}-Warp respectively (bottom of \Cref{tab:results}). \proto{Both models improve on all three robotics benchmarks, and the largest gain}{\modelname-Warp improves on all robotics benchmarks, while \modelname{} improves on \benchmarkreal{} and RGB-Stacking and matches AllTracker on RoboTAP. The largest gain} appears on \benchmarkreal{}, where \modelname{}-Warp reaches 66.0 AJ (+2.1) and is \textbf{the state-of-the-art tracker on every robotics benchmark}. These gains come without sacrificing general video performance, which is maintained or slightly improved on both DAVIS and Kinetics. 

\paragraph{Gains from \benchmarkdata{} transfer to diverse embodiments.} Although \benchmarkdata{} contains only a simulated Franka arm, its gains on \benchmarkreal{} are not limited to Franka footage. \modelname{}-Warp improves over CoWTracker on every video source, including other robot embodiments and human hands. \modelname{}'s gains over AllTracker are smaller and concentrated on robot footage, clearest on DROID and the EgoVerse EVA robot, while it is slightly lower on the human EgoVerse clips. The per-source breakdown and exact numbers are in \Cref{appendix:results_source} (\Cref{fig:per_source_gains} and \Cref{tab:per_source}). 

\subsection{Synthetic Data Ablations}
\label{sec:data_ablation}

\ifallpoints\begin{table*}[t!]
\centering
\caption{\textbf{Data ablations with AllTracker.} ``+'' indicates the additional dataset that was added to Kubric.  ``mix'' is AllTracker's own second-stage mix. Avg.\ is the mean over the five benchmarks. Bold marks the best value within each block.}
\label{tab:data_ablation}
\resizebox{\textwidth}{!}{
\begin{tabular}{l ccc ccc ccc ccc ccc ccc}
\toprule
& \multicolumn{3}{c}{\textbf{\benchmarkreal{}}} & \multicolumn{3}{c}{\textbf{RGB-Stack}} & \multicolumn{3}{c}{\textbf{RoboTAP}} & \multicolumn{3}{c}{\textbf{Kinetics}} & \multicolumn{3}{c}{\textbf{DAVIS}} & \multicolumn{3}{c}{\textbf{Avg.}} \\
\cmidrule(lr){2-4} \cmidrule(lr){5-7} \cmidrule(lr){8-10} \cmidrule(lr){11-13} \cmidrule(lr){14-16} \cmidrule(lr){17-19}
Training data & AJ & $\delta_{\text{vis}}$ & OA & AJ & $\delta_{\text{vis}}$ & OA & AJ & $\delta_{\text{vis}}$ & OA & AJ & $\delta_{\text{vis}}$ & OA & AJ & $\delta_{\text{vis}}$ & OA & AJ & $\delta_{\text{vis}}$ & OA \\
\midrule
\multicolumn{19}{l}{\textit{From AllTracker (Kubric)}} \\
Released checkpoint, no fine-tuning
& 61.4 & 74.5 & 83.9
& 80.7 & 90.1 & 91.7
& 67.5 & 79.3 & 90.8
& 55.8 & 67.9 & 86.9
& 62.8 & 76.2 & 87.2
& 65.6 & 77.6 & 88.1 \\
Kubric only (control)
& 61.5 & 74.6 & 83.9
& 80.6 & 89.7 & 92.1
& 67.5 & 79.3 & 90.7
& 55.9 & 68.1 & 86.9
& 62.4 & 75.9 & 86.7
& 65.6 & 77.5 & 88.1 \\
\quad + PointOdyssey
& 61.2 & \textbf{75.8} & 82.3
& 76.4 & 89.8 & 87.2
& 64.7 & 79.9 & 87.2
& 54.4 & \textbf{68.9} & 83.9
& 63.0 & \textbf{77.0} & 86.6
& 64.0 & 78.3 & 85.4 \\
\quad + \benchmarkdata{}
& 63.0 & \textbf{75.8} & 85.3
& 83.6 & 91.2 & 93.5
& 68.2 & 80.2 & 91.2
& 56.5 & 68.5 & 87.5
& 63.6 & 76.7 & 87.7
& 67.0 & 78.5 & 89.0 \\
\quad + mix + \benchmarkdata{}
& \textbf{63.2} & 75.7 & \textbf{86.1}
& \textbf{84.2} & \textbf{91.6} & \textbf{94.1}
& \textbf{68.9} & \textbf{80.6} & \textbf{92.0}
& \textbf{57.0} & \textbf{68.9} & \textbf{88.4}
& \textbf{63.9} & 76.9 & \textbf{88.1}
& \textbf{67.5} & \textbf{78.7} & \textbf{89.8} \\

\midrule
\multicolumn{19}{l}{\textit{From AllTracker (Kubric+mix), the fully trained model}} \\
Released checkpoint, no fine-tuning
& 62.6 & 75.7 & \textbf{86.9}
& 81.4 & 90.1 & 92.9
& 68.5 & \textbf{80.8} & 91.7
& 56.3 & \textbf{69.0} & \textbf{88.8}
& 64.3 & 77.1 & \textbf{89.5}
& 66.6 & 78.5 & \textbf{90.0} \\
\quad Kubric + \benchmarkdata{} (\modelname{})
& \textbf{63.4} & \textbf{76.1} & 86.2
& \textbf{83.2} & \textbf{90.9} & \textbf{93.5}
& 68.4 & 80.6 & 91.3
& 56.6 & \textbf{69.0} & 87.6
& \textbf{64.6} & \textbf{77.4} & 89.1
& \textbf{67.3} & \textbf{78.8} & 89.5 \\
\quad Kubric + mix + \benchmarkdata{}
& 63.0 & 75.9 & 86.3
& 82.7 & 90.8 & 93.2
& \textbf{68.9} & \textbf{80.8} & \textbf{91.9}
& \textbf{56.8} & \textbf{69.0} & 88.3
& 64.1 & 76.9 & 88.8
& 67.1 & 78.7 & 89.7 \\

\bottomrule
\end{tabular}}
\end{table*}
\else\begin{table*}[t]
\centering
\caption{\textbf{Data ablations with AllTracker.} ``+'' indicates the additional dataset that was added to Kubric.  ``mix'' is AllTracker's own second-stage mix. Avg.\ is the mean over the five benchmarks. Bold marks the best value within each block.}
\label{tab:data_ablation}
\resizebox{\textwidth}{!}{
\begin{tabular}{l ccc ccc ccc ccc ccc ccc}
\toprule
& \multicolumn{3}{c}{\textbf{\benchmarkreal{}}} & \multicolumn{3}{c}{\textbf{RGB-Stack}} & \multicolumn{3}{c}{\textbf{RoboTAP}} & \multicolumn{3}{c}{\textbf{Kinetics}} & \multicolumn{3}{c}{\textbf{DAVIS}} & \multicolumn{3}{c}{\textbf{Avg.}} \\
\cmidrule(lr){2-4} \cmidrule(lr){5-7} \cmidrule(lr){8-10} \cmidrule(lr){11-13} \cmidrule(lr){14-16} \cmidrule(lr){17-19}
Training data & AJ & $\delta_{\text{vis}}$ & OA & AJ & $\delta_{\text{vis}}$ & OA & AJ & $\delta_{\text{vis}}$ & OA & AJ & $\delta_{\text{vis}}$ & OA & AJ & $\delta_{\text{vis}}$ & OA & AJ & $\delta_{\text{vis}}$ & OA \\
\midrule
\multicolumn{19}{l}{\textit{From AllTracker (Kubric)}} \\
Released checkpoint, no fine-tuning
& 61.1 & 74.2 & 83.9
& 80.7 & 90.1 & 91.7
& 71.1 & 82.3 & 92.4
& 59.4 & 71.3 & 89.3
& 62.8 & 76.2 & 87.2
& 67.0 & 78.8 & 88.9 \\
Kubric only (control)
& 61.2 & 74.3 & 83.9
& 80.6 & 89.7 & 92.1
& 71.2 & 82.4 & 92.3
& 59.5 & 71.5 & 89.3
& 62.4 & 75.9 & 86.7
& 67.0 & 78.8 & 88.9 \\
\quad + PointOdyssey
& 61.0 & \textbf{75.6} & 82.3
& 76.4 & 89.8 & 87.2
& 69.7 & 82.9 & 90.1
& 58.3 & \textbf{72.3} & 86.5
& 63.0 & \textbf{77.0} & 86.6
& 65.7 & 79.5 & 86.5 \\
\quad + \benchmarkdata{}
& 62.8 & 75.4 & 85.4
& 83.6 & 91.2 & 93.5
& 71.8 & 82.9 & 92.7
& 60.1 & 72.0 & 89.8
& 63.6 & 76.7 & 87.7
& 68.4 & 79.6 & 89.8 \\
\quad + mix + \benchmarkdata{}
& \textbf{63.0} & 75.4 & \textbf{86.1}
& \textbf{84.2} & \textbf{91.6} & \textbf{94.1}
& \textbf{72.2} & \textbf{83.1} & \textbf{93.1}
& \textbf{60.4} & \textbf{72.3} & \textbf{90.5}
& \textbf{63.9} & 76.9 & \textbf{88.1}
& \textbf{68.8} & \textbf{79.9} & \textbf{90.4} \\

\midrule
\multicolumn{19}{l}{\textit{From AllTracker (Kubric+mix), the fully trained model}} \\
Released checkpoint, no fine-tuning
& 62.3 & 75.4 & \textbf{86.9}
& 81.4 & 90.1 & 92.9
& 71.9 & 83.3 & \textbf{93.1}
& 59.2 & \textbf{72.3} & \textbf{90.4}
& 64.3 & 77.1 & \textbf{89.5}
& 67.8 & 79.6 & \textbf{90.6} \\
\quad Kubric + \benchmarkdata{} (\modelname{})
& \textbf{63.2} & \textbf{75.9} & 86.4
& \textbf{83.2} & \textbf{90.9} & \textbf{93.5}
& 72.2 & 83.4 & 92.7
& \textbf{60.0} & \textbf{72.3} & 89.8
& \textbf{64.6} & \textbf{77.4} & 89.1
& \textbf{68.6} & \textbf{80.0} & 90.3 \\
\quad Kubric + mix + \benchmarkdata{}
& 62.8 & 75.6 & 86.4
& 82.7 & 90.8 & 93.2
& \textbf{72.5} & \textbf{83.6} & \textbf{93.1}
& \textbf{60.0} & \textbf{72.3} & 90.3
& 64.1 & 76.9 & 88.8
& 68.4 & 79.8 & 90.4 \\

\bottomrule
\end{tabular}}
\end{table*}

\fi

\paragraph{Experimental setup.}
We study the effect of synthetic training data through ablations on AllTracker~\citep{harley2025alltracker}, which at 17M parameters is about 60$\times$ smaller than CoWTracker and makes repeated fine-tuning runs practical. AllTracker is trained in two stages, first on Kubric and then for 400K steps on a mixture of optical-flow and point-tracking datasets (Kubric+mix). Starting from the released Kubric checkpoint, we replace the second stage with 10k fine-tuning steps at a learning rate of $10^{-5}$, mixing each candidate dataset equally with Kubric as replay data to preserve general tracking. We compare against a Kubric-only continuation (\Cref{tab:data_ablation}) and study the mixing ratio separately below. 

\paragraph{Manipulation-focused synthetic data improves tracking on both robotic and general videos.}
We first ask whether synthetic data aligned with manipulation helps at all, comparing \benchmarkdata{} with PointOdyssey, a general-purpose synthetic dataset, and with the Kubric-only control under the same fine-tuning stage (\Cref{tab:data_ablation}). Fine-tuning with \benchmarkdata{} improves AJ over the control on all five benchmarks, by 1.5 on \benchmarkreal{}, 3.0 on RGB-Stacking, 1.2 on DAVIS, \proto{and 0.6 each on RoboTAP and Kinetics}{0.7 on RoboTAP, and 0.6 on Kinetics}. These gains are notable because \benchmarkdata{} contains only synthetic interactions from a single robot and task, yet transfers to diverse real-world manipulation and general videos. In contrast, fine-tuning with PointOdyssey does not yield comparable improvements, suggesting that the types of motions represented in synthetic training data matter for transfer.

\begin{figure}[t!]
    \centering
\ifallpoints
    \begin{minipage}[t]{0.485\linewidth}
        \centering
        \includegraphics[width=\linewidth]{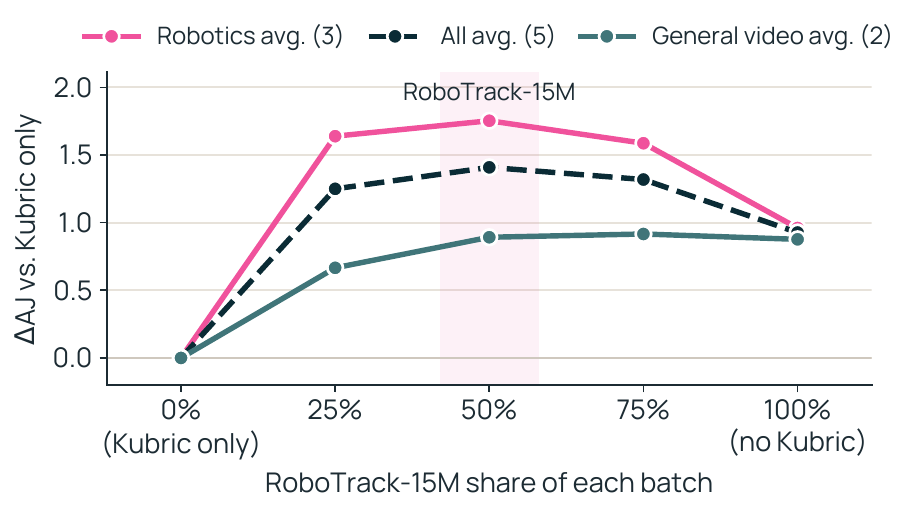}
        \caption{\textbf{Effect of the \benchmarkdata{} share in each batch.} $\Delta$AJ is relative to the Kubric-only second stage. We use the 50\% mix (shaded), which gives the best average AJ across all benchmarks (\proto{68.4, vs.\ 67.0}{67.0, vs.\ 65.6} for Kubric only).}
        \label{fig:data_mix}
    \end{minipage}\hfill
    \begin{minipage}[t]{0.485\linewidth}
        \centering
        \includegraphics[width=\linewidth]{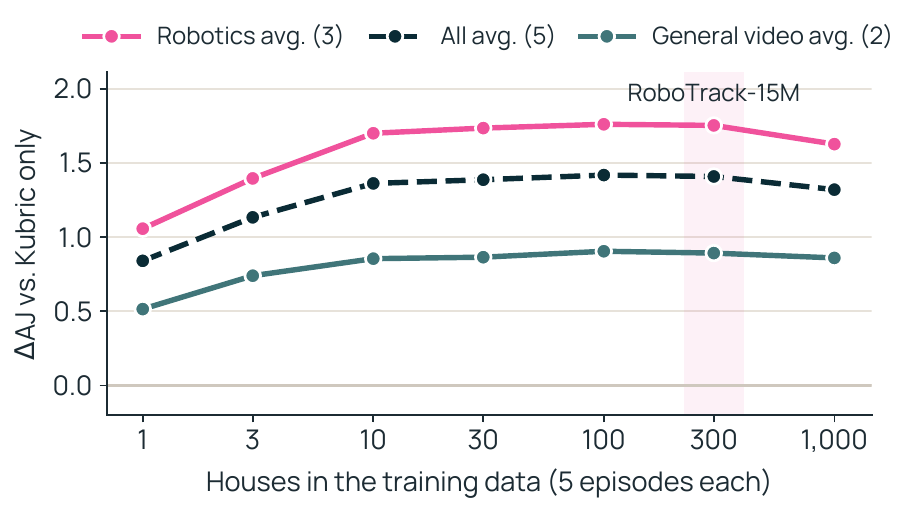}
        \caption{\textbf{Effect of the number of houses in the training data.} $\Delta$AJ is relative to the Kubric-only second stage. The 50\% recipe trained on nested subsets of 1 to 1,000 houses (5 episodes each); gains level off at about 10 houses.} 
        \label{fig:house_scaling}
    \end{minipage}
\else
    \includegraphics[width=0.6\linewidth]{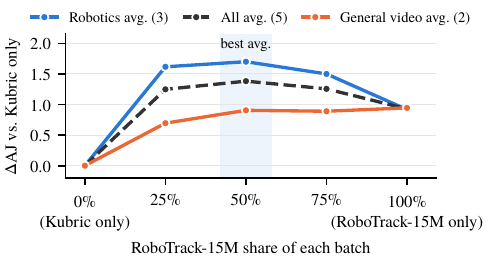}
    \caption{\textbf{Effect of the \benchmarkdata{} share in each batch.} $\Delta$AJ is relative to the Kubric-only second stage. We use the 50\% mix (shaded), which gives the best average AJ across all benchmarks (\proto{68.4, vs.\ 67.0}{67.0, vs.\ 65.6} for Kubric only).}
    \label{fig:data_mix}
\fi
\end{figure}

\paragraph{A short fine-tune on \benchmarkdata{} can match substantially longer training on existing tracking datasets.} We next ask how this short stage compares with AllTracker’s full second stage on its mixture of existing tracking datasets. Despite using only 10k second-stage training steps, versus AllTracker's original 400K, our model outperforms the released fully trained AllTracker on \benchmarkreal{} (63.0 vs.\ 62.6 AJ)\proto{, RGB-Stacking (83.6 vs.\ 81.4), and Kinetics (60.1 vs.\ 59.2). Performance is comparable on RoboTAP (71.8 vs.\ 71.9) and DAVIS (63.6 vs.\ 64.3).}{ and RGB-Stacking (83.6 vs.\ 81.4), is comparable on Kinetics (56.5 vs.\ 56.3) and RoboTAP (68.2 vs.\ 68.5), and is slightly lower on DAVIS (63.6 vs.\ 64.3).} The datasets are also complementary: applying the same fine-tune to the fully trained AllTracker improves \proto{all five benchmarks}{\benchmarkreal{} and RGB-Stacking and is comparable on the other three benchmarks}, reaching 63.4 AJ on \benchmarkreal{}.

\paragraph{A balanced mixture of manipulation data and Kubric gives the best overall performance.}
We vary the proportion of \benchmarkdata{} in each training batch from 0\% to 100\%, with Kubric making up the remainder (\Cref{fig:data_mix}). Most of the improvement on robotic benchmarks occurs with the first 25\% of manipulation data, increasing average AJ by \proto{1.6}{1.6} points before performance levels off. Interestingly, training on manipulation data alone reduces the robotics gain to \proto{0.9}{1.0} AJ, suggesting that mixing with diverse sources such as Kubric is crucial. The 50\% mixture achieves the highest average AJ across all five benchmarks (\proto{68.4 vs.\ 67.0}{67.0 vs.\ 65.6} for Kubric alone), which we use to train \modelname{}.

\paragraph{Scaling simulated scenes improves performance with diminishing returns.} We next ask how performance changes with the number of scenes in \benchmarkdata{}, training the same 50\% mixture on nested subsets of 1 to 1,000 houses with five episodes each (\Cref{fig:house_scaling}). Gains grow as houses are added, reaching +1.7 AJ on robotics benchmarks at 10 houses and +1.8 at both 100 and 300 houses, beyond which additional houses have a degrading effect (+1.6 at 1,000). We hypothesize that, because every house contributes the same pick-up task, further gains will come from adding tasks and embodiments rather than more scenes. We therefore use the 300-house set as \benchmarkdata{}.

\begin{figure*}[ht!]
  \centering
  \setlength{\tabcolsep}{0pt}
  \begin{tabular}{@{}c@{\hspace{0.03\textwidth}}c@{}}
    \includegraphics[width=0.47\textwidth]{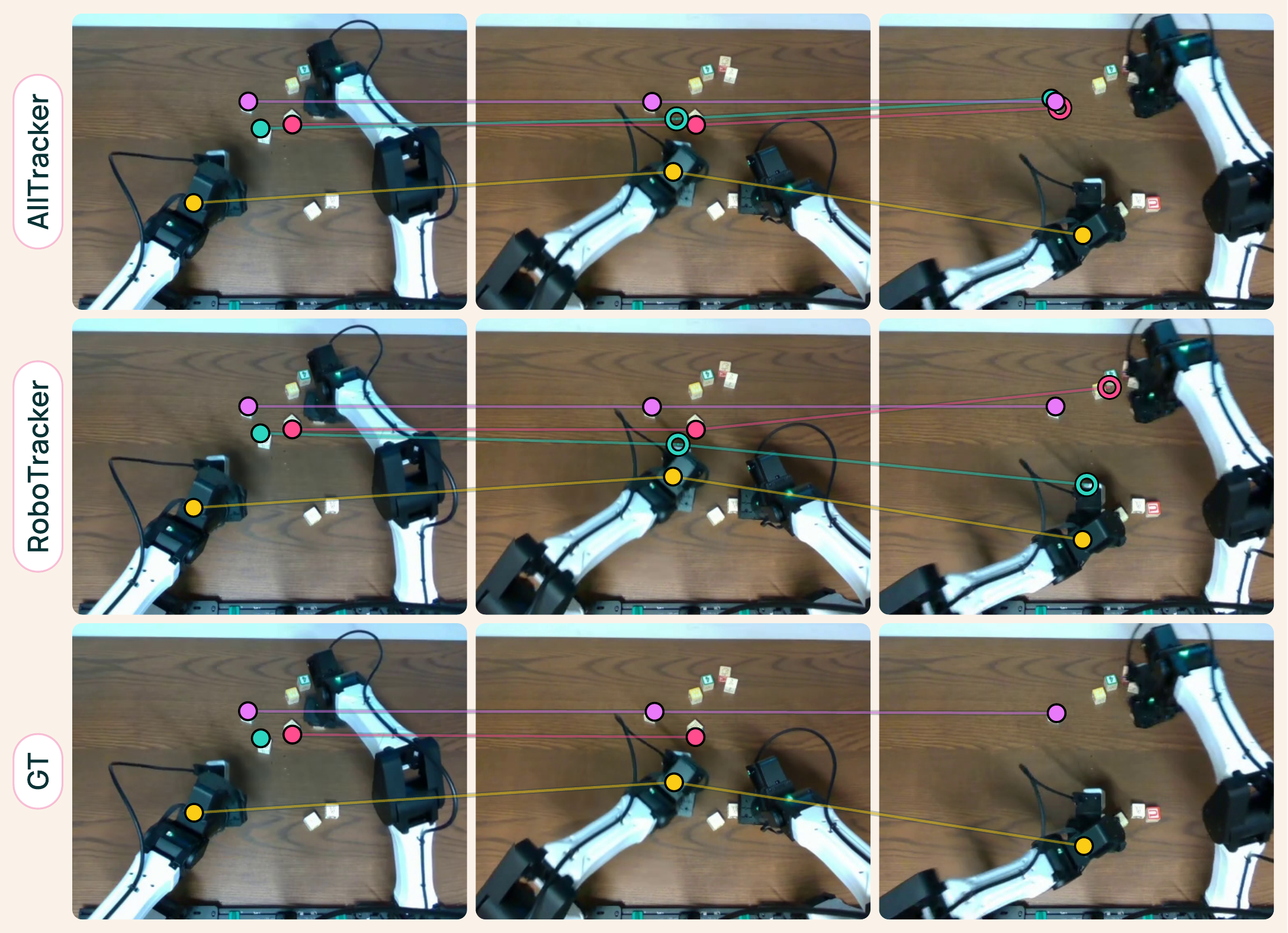} &
    \includegraphics[width=0.47\textwidth]{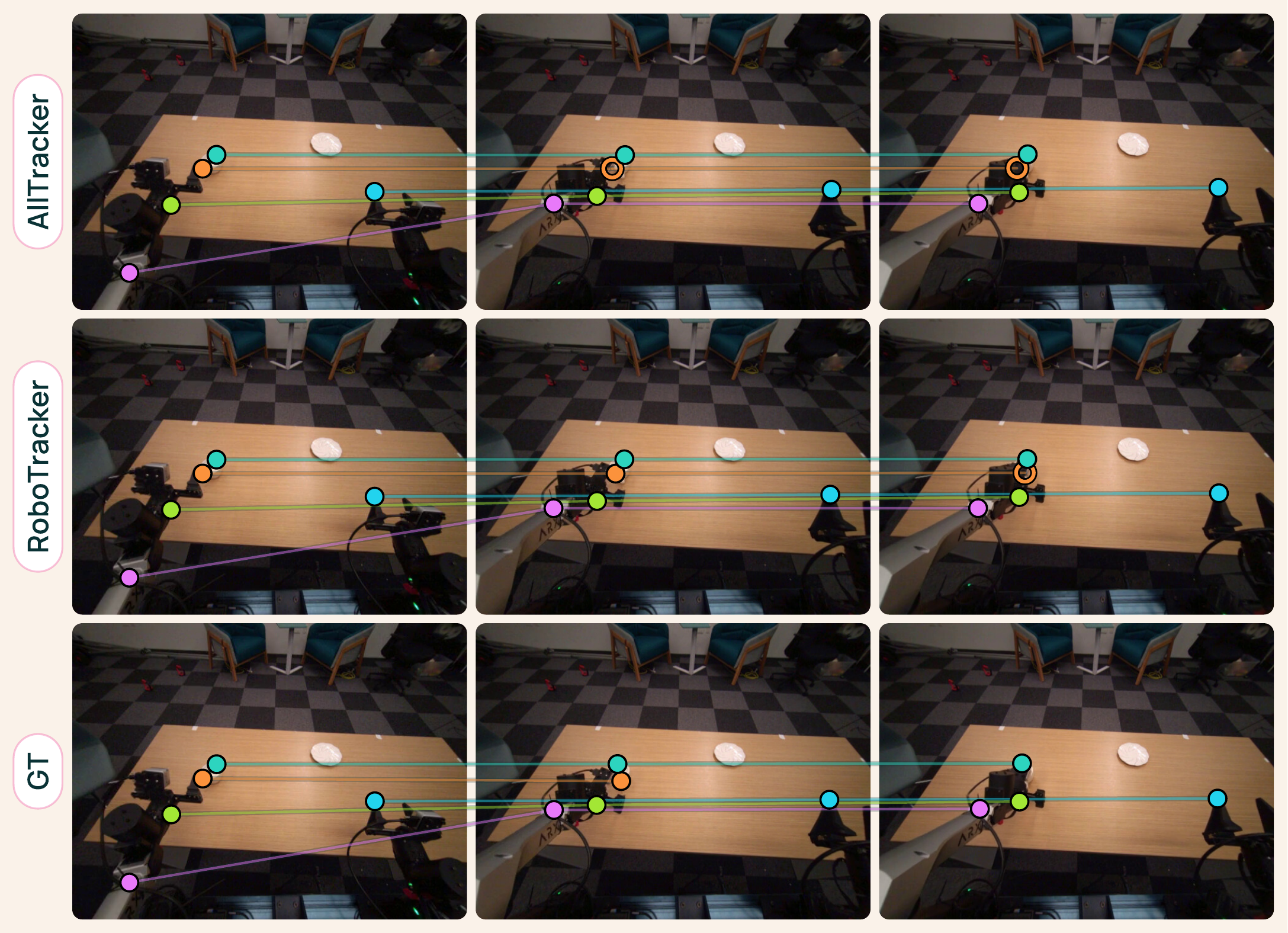} \\
    \small (a) MolmoAct2 & \small (b) EgoVerse \\[6pt]
    \includegraphics[width=0.47\textwidth]{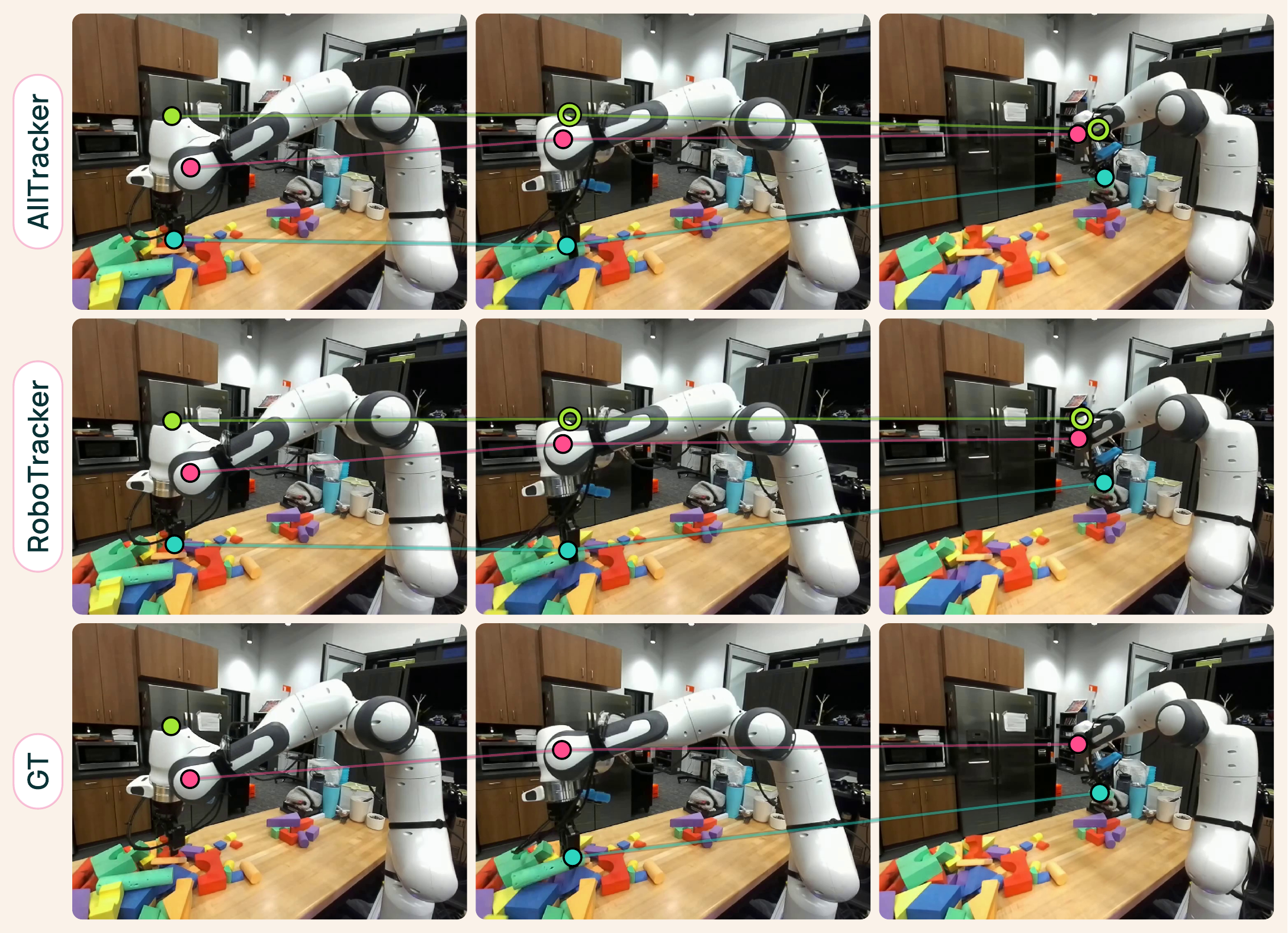} &
    \includegraphics[width=0.47\textwidth]{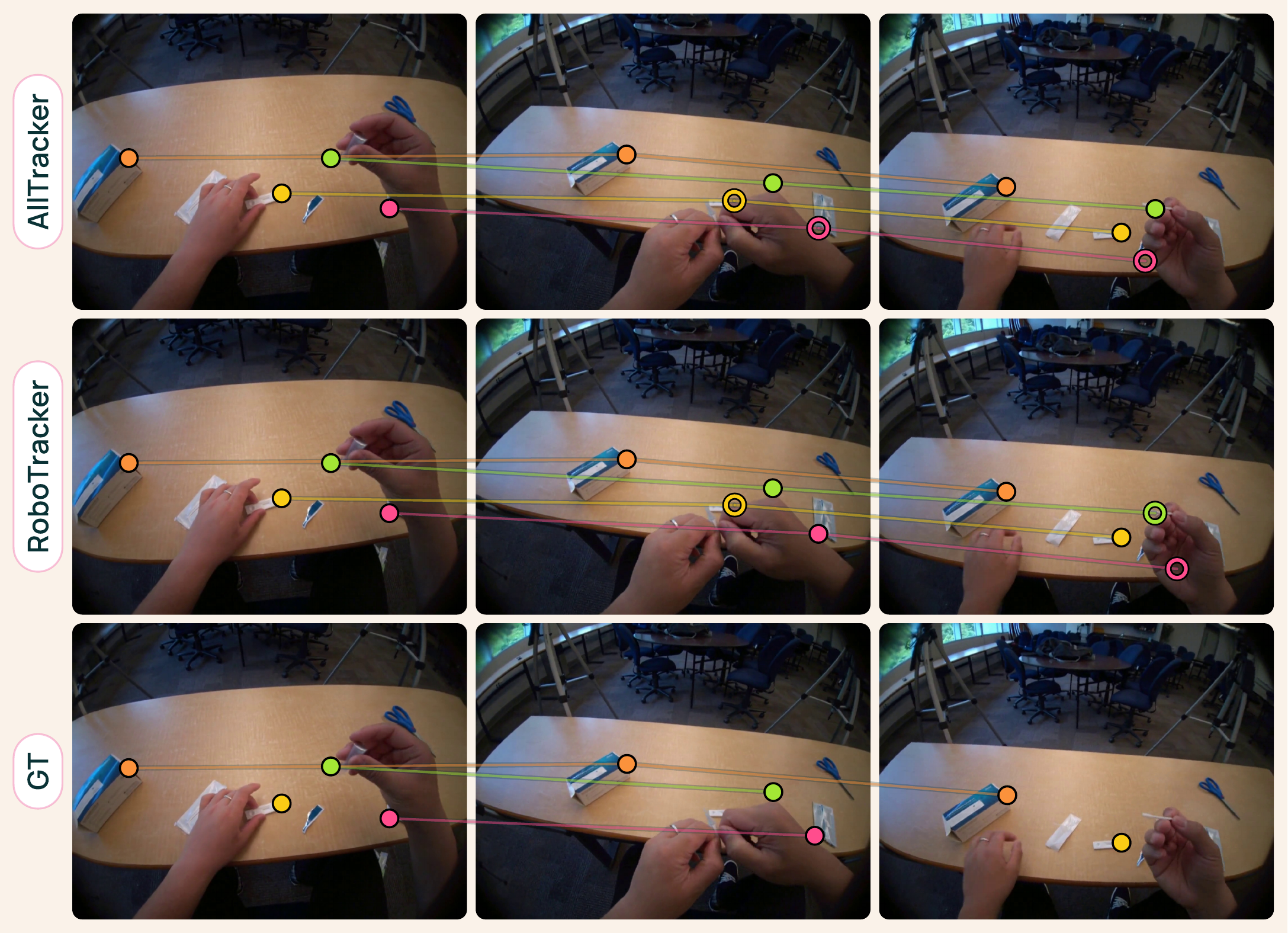} \\
    \small (c) DROID & \small (d) Ego-Exo4D
  \end{tabular}
  \caption{\textbf{AllTracker vs.\ RoboTracker on RoboTrack-Bench.} Hollow points indicate that the model predicts the point is occluded in that frame. Overall, RoboTracker keeps points on the correct objects during manipulation and handles occlusion better.}
  \label{fig:qual_compare}
\end{figure*}

\subsection{Qualitative Results}
\label{sec:qualitative}
\Cref{fig:qual_compare} compares AllTracker~\cite{harley2025alltracker} with \modelname{} on one \benchmarkreal{} clip per source. The benefit of fine-tuning on \benchmarkdata{} is clearly visible across all four sources. On MolmoAct2 (a), \modelname{} keeps the points on the correct blocks, including blocks that become occluded. On EgoVerse (b), it places the point on the manipulated cup more precisely. On DROID (c), it correctly predicts which points are occluded, while AllTracker still draws hidden points. On Ego-Exo4D (d), it better localizes the occluded point behind the right hand, drawn in pink. 

\Cref{fig:warp_qual} shows the same comparison between CoWTracker and \modelnamewarp{}. Overall, \modelnamewarp{} stays on the points and predicts their visibility more accurately, especially on the moving robot and hands. On MolmoAct2 (a), it correctly predicts which points on the moving gripper are visible, while CoWTracker marks a visible point on the gripper as occluded and draws a hidden point next to it as visible. On EgoVerse (b), it follows the two points on the blanket as the person pulls it to the edge of the table, while CoWTracker first marks them as occluded and then jumps to the person's other hand. On DROID (c), it stays on the point on the arm as the arm turns and marks the point near the gripper as occluded once it is hidden, while CoWTracker drifts off the first point and keeps drawing the second after it disappears. On Ego-Exo4D (d), both models follow the background points as the camera moves, but \modelnamewarp{} places the point on the object in the person's hands more precisely.

We show failure cases of our models in \Cref{appendix:failure_cases}.

\section{Robotics Application: Wet Lab Automation from a Single Demonstration}
\label{sec:wetlab}

\begin{figure}[!t]
    \centering
    \includegraphics[width=\linewidth]{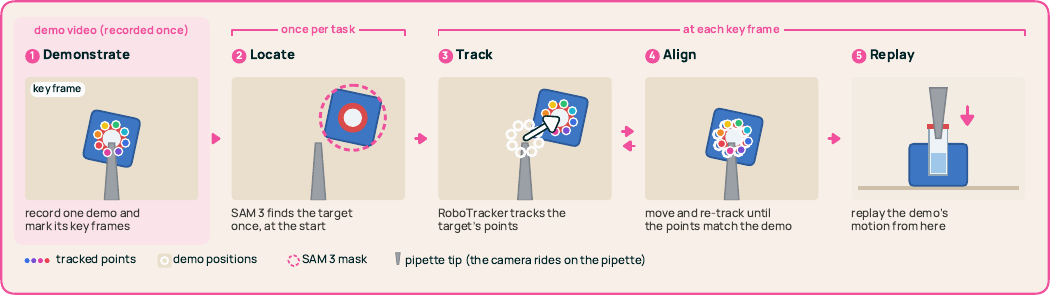}
    \caption{\textbf{Guiding the robot with point tracks from a single demonstration.} We record one demonstration and mark its key frames (1). At test time, SAM3 locates the target once (2). At each key frame, \modelname{} tracks the target's points (3), the robot moves and re-tracks until the points match their demonstrated positions (4), and then replays the demonstrated motion (5). The camera is mounted on the pipette.}
    \label{fig:wetlab_method}
\end{figure}

As a real-world application, we consider wet lab automation, which we solve by using correspondences to align with a demonstration sequence, building on FlowControl~\citep{argus2020flowcontrol} and similar to RoboTAP~\citep{vecerik2024robotap}. We chose the common task of pipetting a liquid into a vial; this task is a multi-step sequence that requires high-precision control, in which the robot grasps the pipette, draws liquid from the first vial, dispenses it into the second, and places the pipette back. \Cref{fig:wetlab_method} illustrates our pipeline: at each key frame of the demonstration, the robot tracks points on the target and moves until they match their demonstrated positions, then replays the demonstrated motion. 

\subsection{Task and Setup}

\paragraph{Task. }
The robot picks up the pipette, transfers liquid between two vials, and places the pipette back on the holder. The positions of the pipette and vials change between trials. The task has four vision-guided steps: grasp the pipette, move to vial 1 and aspirate, move to vial 2 and dispense, and return the pipette to its holder. Steps 1 and 4 are guided by the wrist camera while steps 2 and 3 are guided by the pipette camera. A trial counts as a task success only if all four steps succeed. 

\paragraph{Setup. }
We use one arm of a bimanual I2RT YAM robot with a parallel gripper. Two Intel RealSense D405 RGB-D cameras provide observations at $848\times480$ and 15 fps: one on the wrist, and one mounted on the pipette looking down along its tip. The pipette is an electronic pipette that the robot picks up from a holder and that is triggered over the network. 

\FloatBarrier

\subsection{Alignment from One Demonstration}

\paragraph{One-shot imitation by alignment.}
Similar to RoboTAP~\citep{vecerik2024robotap}, the robot repeats a single demonstration. The demonstration is recorded once by teleoperation and divided into steps, each with a keyframe image and a mask of the object that the step acts on (the pipette holder, or the colored ring around a vial). At test time, for each step the robot moves to the keyframe pose, estimates how far the object has moved relative to the demonstration by comparing the keyframe with the live image, shifts the end-effector by that offset, and then replays the demonstrated motion for the step. Since the replay is open-loop, the accuracy of the estimated offset determines whether the pipette tip enters the vial.

\paragraph{Alignment methods. }
We compare three ways of estimating the offset. All share the same demonstration, controller, and replay, and all are given the same budget of one coarse move followed by three corrections. 

\begin{itemize}
    \item \textbf{RAFT (flow)}: optical flow~\citep{teed2020raft} from the keyframe to the live image is read at the object's mask pixels, lifted to 3D with depth, and a rigid motion is fit with RANSAC. 
    \item \textbf{SAM3 (centroid)}: the object is segmented in the live image from a text prompt with SAM3~\citep{carion2025sam3segmentconcepts}, and the end-effector moves so that the centroid of the mask matches the centroid in the demonstration. Every correction uses a new detection.
    \item \textbf{\modelname{} (track)}: the same fit as for flow, with correspondences from our tracker. Each vial is segmented once with SAM3 when the pipette is picked up, and the tracker carries the points forward through the rest of the task. At each correction, the tracker receives the reference frame, up to 14 frames of the video since then, and the live frame, and matches the points of the demonstration mask to the live image. Points that the tracker predicts as occluded are kept in the fit, so the tracker can align when the target is partially hidden. 
\end{itemize}

\subsection{Results}

\paragraph{Protocol. }
We evaluate each method on 20 layouts, 10 with occlusion and 10 without (\Cref{tab:real_world}). The locations of vials were randomly shifted between layouts. Occlusions occurred in the middle of the task to test each method's ability to track under intermittent occlusion. Every method completed the grasp and place steps in all trials, so the comparison focuses on the two vial alignment steps.

\begin{figure}[!t]
    \centering
    \includegraphics[width=\textwidth]{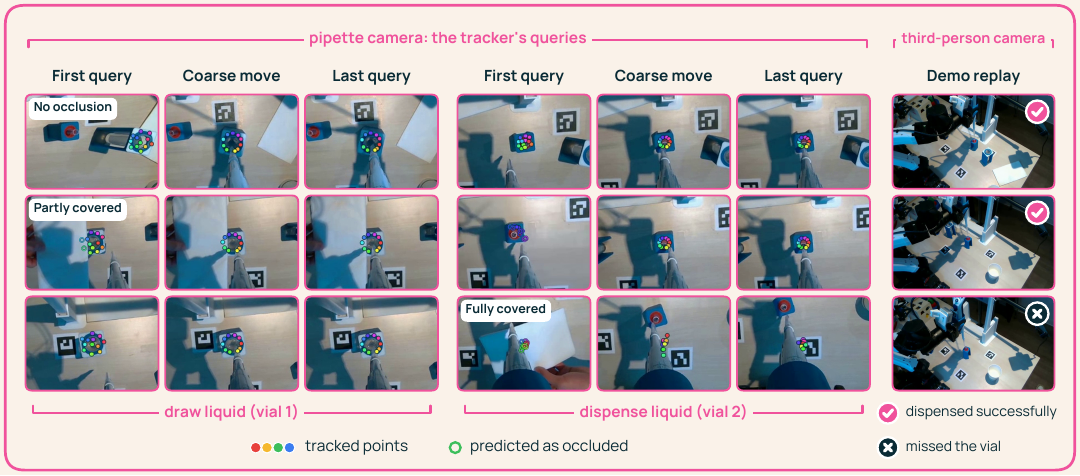}
    \caption{\textbf{Tracker queries during the pipette task.} Each row is one complete trial of \modelname{} (track). The first six columns show the pipette camera at the first query, after the coarse move, and at the last query for the two vials. The last column shows a third-person camera while the demonstrated motion is replayed over vial 2. Filled dots are tracked points on the ring around the vial, and hollow rings are points that the tracker predicts as occluded.}
    \label{fig:robot_qual}
\end{figure}

\begin{table}[!ht]
\centering
\caption{\textbf{Real-world pipette transfer.} Successes per step over 20 total layouts, one trial per method per layout, with vial positions changed between layouts. Every method grasped and placed the pipette in every trial successfully, so we report only the draw and dispense steps. \emph{Success} counts trials where both succeeded. }
\label{tab:real_world}
\resizebox{0.8\columnwidth}{!}{
\begin{tabular}{l ccc ccc}
\toprule
& \multicolumn{3}{c}{\textbf{No occlusion}} & \multicolumn{3}{c}{\textbf{Occlusion}} \\
\cmidrule(lr){2-4} \cmidrule(lr){5-7}
Method & Draw & Dispense & Success & Draw & Dispense & Success \\
\midrule
RAFT (flow)        & 5/10            & 5/10              & 3/10              & \underline{4/10} & 2/10              & \underline{2/10} \\
SAM3 (centroid)    & \textbf{10/10}  & \underline{7/10}  & \underline{7/10}  & \underline{4/10} & \underline{5/10}  & 1/10 \\
\modelname{} (track)  & \textbf{10/10}  & \textbf{10/10}    & \textbf{10/10}    & \textbf{9/10}    & \textbf{8/10}     & \textbf{7/10} \\
\bottomrule
\end{tabular}}
\end{table}

We evaluate our results quantitatively in~\Cref{tab:real_world}, and find that our approach is more robust to large displacements compared to optical flow-based correspondences and more robust to occlusion compared to centroid-based alignment.

\paragraph{Results and failure analysis. }

The three approaches fail in different ways, and the differences show what tracking provides. 

\emph{Optical flow} estimates correspondences directly from the demonstration keyframe to the live image. This works well when the vials are close to their positions in the demonstration, but fails when they are further away. As a result, flow only succeeds on half of the vial alignments even with no occlusions. 

\emph{Centroid alignment} using SAM3 ~\citep{carion2025sam3segmentconcepts} masks is capable of handling both large and small displacements. However, since every look depends on a fresh detection with no previous knowledge, SAM3 can become confused by a duplicate or similar object, which is what caused the failures under the no occlusion case. Under occlusion, SAM3 might not find the object, or even if it does, the difference in the mask's shape causes centroid alignment to fail. These factors together cause SAM3 to get a score of 7 under no occlusion and 1 with occlusions. 

\emph {Tracking} identifies the target once and tracks it throughout the video. This allows the method to handle large displacements and track the target accurately under occlusion, mitigating the weaknesses of both previous methods. This lets tracking achieve strong performance in both no occlusion and occlusion, scoring 10 and 7 respectively. Tracking still fails under extended periods of occlusion or when the target is completely occluded because tracking accuracy decreases under occlusion.

\paragraph{Qualitative examples. }
\Cref{fig:robot_qual} shows the tracker's queries over three complete trials of \modelname{} (track), at vial 1, where the robot draws the liquid, and at vial 2, where it dispenses it. Without occlusion (top), the tracked points lie on the ring around each vial at the first query and stay on it as the robot moves, until the ring sits under the pipette tip at the last query. The replayed motion then lowers the tip into the vial. When a sheet of paper covers part of the ring around vial 1 (middle), the tracker predicts the points under the paper as occluded but still estimates their positions, and the alignment succeeds at both vials. In the failure (bottom), the draw at vial 1 succeeds, but the paper covers vial 2 completely at the first query and the tracker predicts every point as occluded. After the paper is removed, the points stay on the table beside the pipette tip instead of returning to the vial, so the pipette misses it.

\section{Conclusion}
\label{sec:conclusion}
We introduce \benchmark{}, pairing the real-world benchmark \benchmarkreal{} with \benchmarkdata{}, synthetic training data generated in MolmoSpaces. Strong results on RGB-Stacking or RoboTAP do not necessarily transfer to \benchmarkreal{}, suggesting that current evaluations overestimate real-world readiness. Fine-tuning on \benchmarkdata{} improves every robotics benchmark for \modelname{}-Warp and \benchmarkreal{} for both models. On a real robot, \modelname{} successfully guides pipette handling in a wet lab. Future work includes better occlusion handling and richer simulated interactions.

\clearpage
\bibliographystyle{abbrvnat}
\bibliography{main}

\clearpage
\appendix
\crefalias{section}{appendix}
\crefalias{subsection}{appendix}
\crefname{appendix}{Appendix}{Appendices}
\Crefname{appendix}{Appendix}{Appendices}
\section*{Appendix}

The appendix includes the following sections:
\begin{itemize} 
\itemsep0em 
    \item \S\ref{appendix:benchmark_stats} - \benchmarkreal{} Statistics
    \item \S\ref{appendix:benchmark_annotation} - \benchmarkreal{} Annotation Details
    \item \S\ref{appendix:benchmark_data} - \benchmarkdata{} details
    \item \S\ref{appendix:cowtracker_control} - Training CoWTracker
    \item \S\ref{appendix:results_source} - Results by Video Source
    \item \S\ref{appendix:qual_examples} - Qualitative Examples 
    \item \S\ref{appendix:limitation} - Limitations
    \item \S\ref{appendix:broader_impact} - Broader Impact
    \item \S\ref{appendix:licenses} - Licenses
\end{itemize}

\section{\benchmarkreal{} Statistics}
\label{appendix:benchmark_stats}

\begin{figure*}[ht!]
    \centering
    \includegraphics[width=\textwidth]{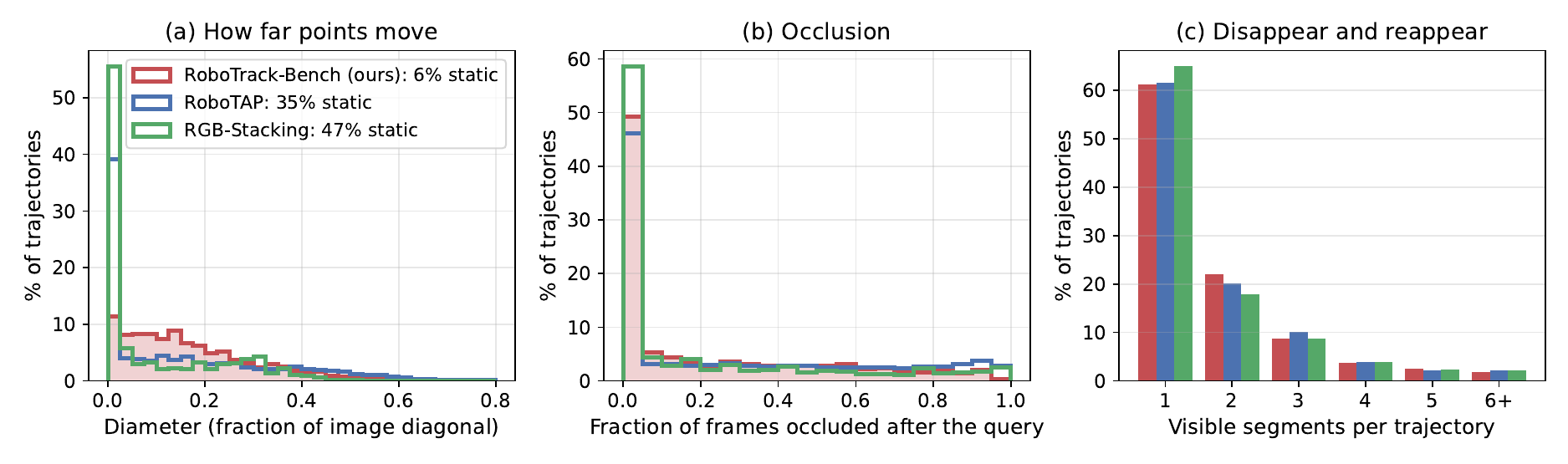}
    \caption{\textbf{Track statistics of \benchmarkreal{} compared with RoboTAP and RGB-Stacking.} (a) Trajectory diameter, the maximum distance between the positions of a point over time, as a fraction of the image diagonal: only 6\% of points in \benchmarkreal{} stay static (moving less than 1\% of the diagonal), compared with 35\% in RoboTAP and 47\% in RGB-Stacking. (b) Fraction of frames in which a point is occluded after its query frame. (c) Number of visible segments per trajectory, i.e., contiguous visible sections separated by occlusions; 39\% of points in \benchmarkreal{} disappear and reappear at least once. Each benchmark is normalized by its own number of trajectories.}
    \label{fig:droid_stats}
\end{figure*}

Figure~\ref{fig:droid_stats} compares the track statistics of \benchmarkreal{} with robotics tracking benchmarks such as RoboTAP and RGB-Stacking. Clips in \benchmarkreal{} average 223 frames (26 to 785) and 8.2 tracks. 79\% of tracks are queried on the first frame, with the rest queried when the point first becomes visible. 43\% of tracks stay visible for their whole length, a track is occluded for 22\% of its frames on average, and 39\% of tracks disappear and reappear at least once. The median trajectory diameter is 14\% of the image diagonal, in line with our aim of capturing intricate arm movements rather than large camera sweeps. 
\Cref{fig:benchmark_examples} shows qualitative examples from \benchmarkreal{}. Overall, we observe that point tracks are accurate and consistent across videos. We find that annotations with fewer refinement points are often smoother, indicating that the optical flow tracking assistant is very helpful for obtaining high-quality point tracks.

\section{\benchmarkreal{} Annotation Details}
\label{appendix:benchmark_annotation}
\subsection{Track assist algorithm.}
\label{appendix:track_assist}
Our track assist algorithm closely follows TAP-Vid~\citep{tapvid}, with the main difference being the optical flow model. For every video, we compute optical flow using SEA-RAFT~\citep{wang2024sea}, which is used to generate assisted tracks for the annotator. The assisted tracks are usually accurate over a few frames, but cannot handle occlusions and tend to drift over time; annotators are responsible for marking occlusions and adding refinement points when the track drifts.

When an annotator adds a point $p_s$ on frame $s$, we use optical flow to propagate that point from one frame to the next for the entire video. When the annotator chooses a second point $p_t$ on frame $t$, we use Dijkstra's algorithm to find the shortest path through the optical flow from $p_s$ to $p_t$. Annotators are also always given the option to linearly interpolate between the two points instead, which we found helpful for certain points.

\subsection{Annotation Interface}
\label{appendix:annotation_ui}

\begin{figure*}[t]
    \centering
    \includegraphics[width=\textwidth]{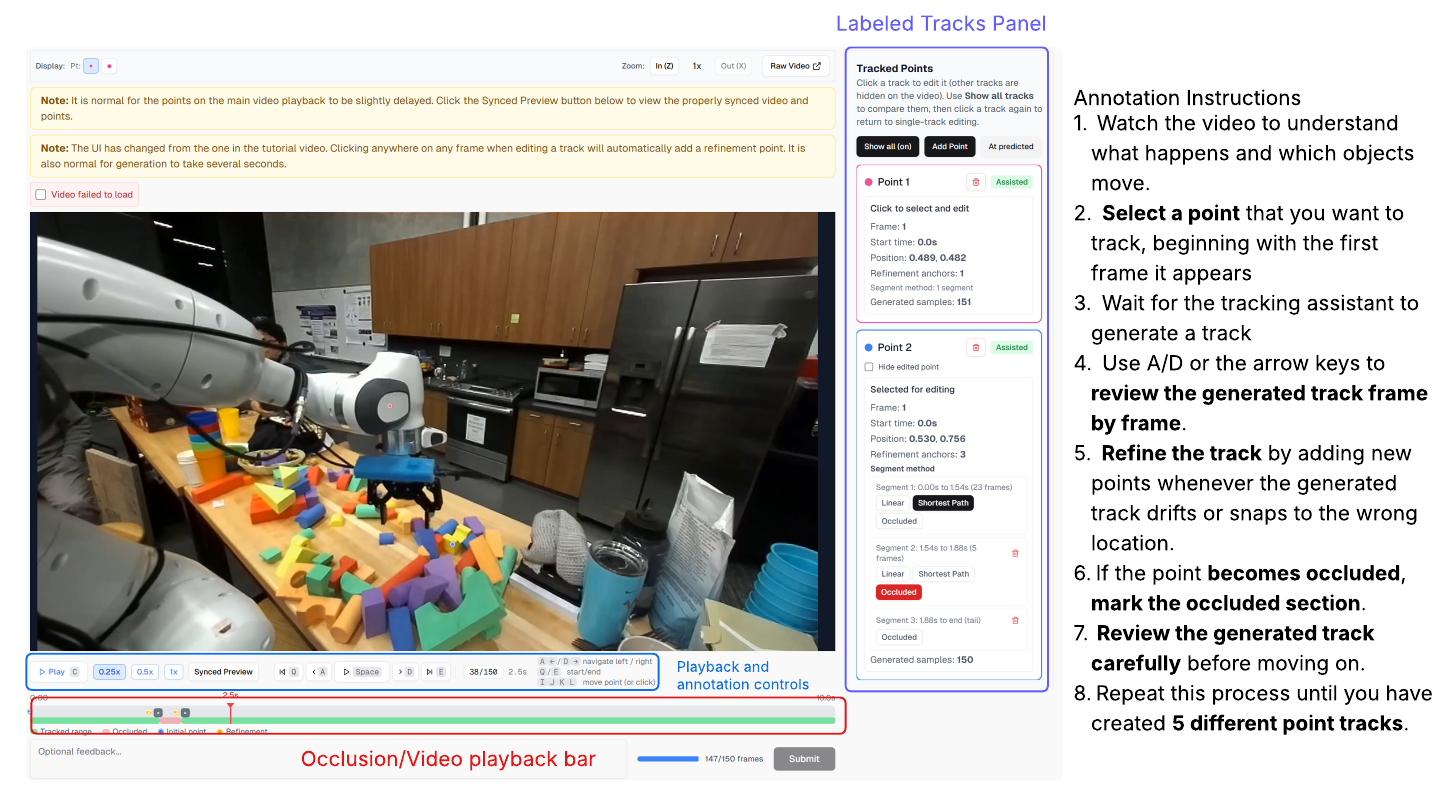}
    \caption{\textbf{Point annotation interface and annotation instructions for \benchmarkreal{}.} The interface consists of two main sections: the video on the left, with frame-by-frame seeking, and the labeled tracks on the right.}
    \label{fig:annotation_ui}
\end{figure*}

\Cref{fig:annotation_ui} shows the annotation interface and the instructions given to annotators for labeling \benchmarkreal{} (\Cref{sec:real}). The left side is the main video display, where annotators click on the video to add points and use the playback controls to seek through the video. The right side lists the currently tracked points and relevant information for each. Each point's track is divided into time segments by the refinement points the annotator selected; for each segment, the annotator can choose an interpolation method or mark the segment as occluded.

\section{\benchmarkdata{} Details}
\label{appendix:benchmark_data}

We provide additional statistics of \benchmarkdata{} and describe the strategies we use to create the final point tracks used for training.

\begin{figure*}[t]
    \centering
    \includegraphics[width=\textwidth]{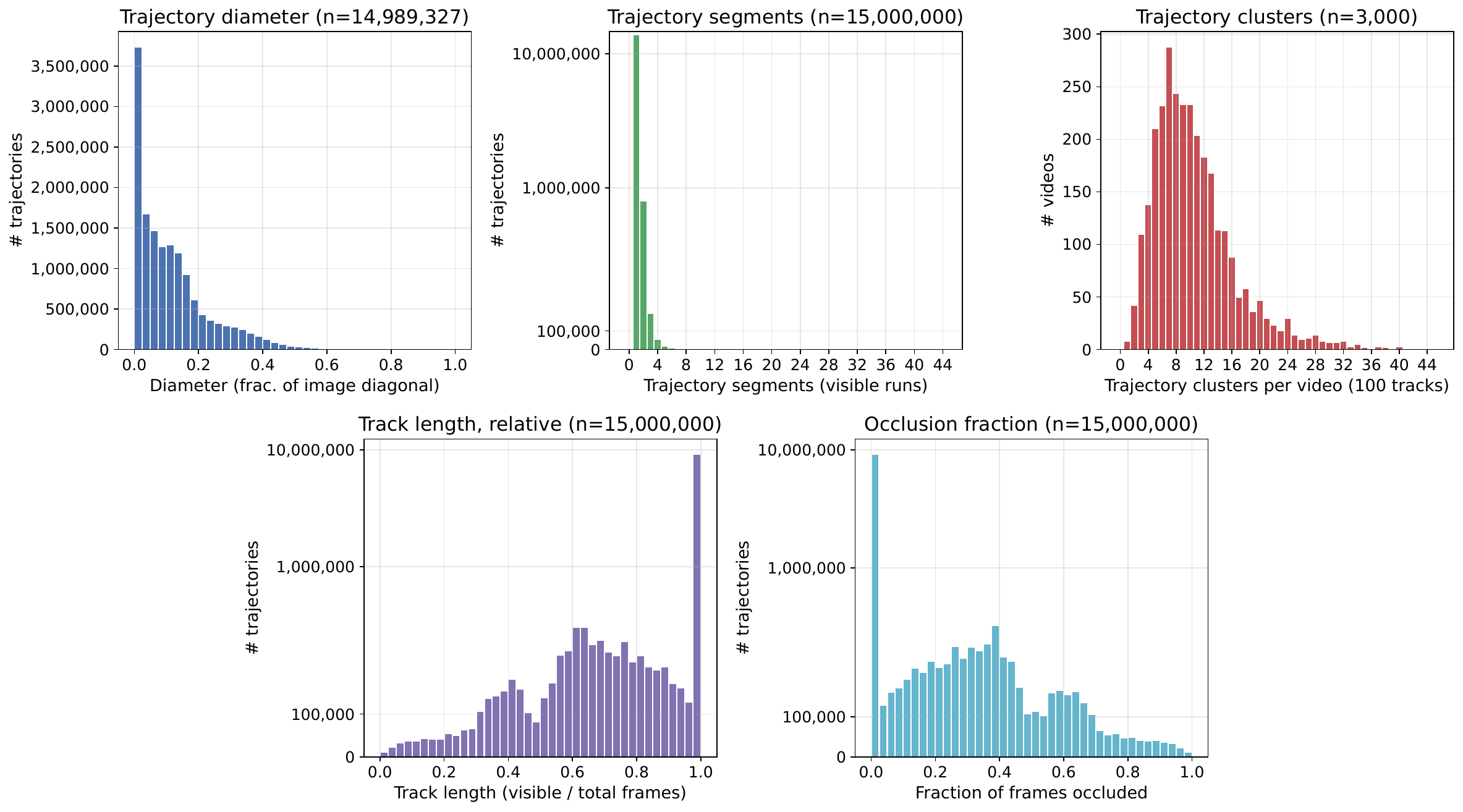}
    \caption{\textbf{Statistics of tracks in \benchmarkdata{}.} ``Diameter'' refers to the maximum distance between the positions of a point over time. Trajectory segments refer to the number of contiguous visible sections of a trajectory, with breaks due to occlusion. Trajectory clusters count the number of motion-distinct groups per video; higher values mean more diverse motion.}
    \label{fig:molmospaces_stats}
\end{figure*}

\paragraph{Dataset Statistics} 
Table~\ref{tab:data_stats} summarizes the composition of \benchmarkdata{}. \Cref{fig:molmospaces_stats} further reports the distribution of track-level statistics in \benchmarkdata{}, log scaled. 

\paragraph{Obtaining point tracks.}
We now describe in more detail how point tracks are obtained from the MolmoSpaces simulation (\Cref{sec:sim}). Given a pixel $(u, v)$ in the query frame with rendered depth $d$, we unproject it to world coordinates:
\begin{equation}
\mathbf{p}^{\mathrm{world}} = \mathbf{R}_{\mathrm{cam}} \begin{pmatrix} (u - c_x)\, d / f \\ (v - c_y)\, d / f \\ d \end{pmatrix} + \mathbf{t}_{\mathrm{cam}},
\label{eq:unproject}
\end{equation}
where $f$ is the focal length, $(c_x, c_y)$ is the principal point (image center), and $\mathbf{R}_{\mathrm{cam}}$ and $\mathbf{t}_{\mathrm{cam}}$ are the camera's rotation and position in world coordinates. We then express the point in the local frame of its parent body, with rotation $\mathbf{R}_{\mathrm{body}}$ and position $\mathbf{t}_{\mathrm{body}}$:
\begin{equation}
\mathbf{p}^{\mathrm{local}} = \mathbf{R}_{\mathrm{body}}^\top \left(\mathbf{p}^{\mathrm{world}} - \mathbf{t}_{\mathrm{body}}\right).
\label{eq:body_local}
\end{equation}
At each subsequent frame $t$, the world position is recovered from the body pose as $\mathbf{p}^{\mathrm{world}}_{t} = \mathbf{R}_{\mathrm{body},t}\, \mathbf{p}^{\mathrm{local}} + \mathbf{t}_{\mathrm{body},t}$, transformed into the camera frame as $\mathbf{p}^{\mathrm{cam}}_{t} = \mathbf{R}_{\mathrm{cam},t}^\top \left(\mathbf{p}^{\mathrm{world}}_{t} - \mathbf{t}_{\mathrm{cam},t}\right)$, and projected with the pinhole model:
\begin{equation}
u_{t} = f \frac{p^{\mathrm{cam}}_{t,x}}{p^{\mathrm{cam}}_{t,z}} + c_x, \qquad
v_{t} = f \frac{p^{\mathrm{cam}}_{t,y}}{p^{\mathrm{cam}}_{t,z}} + c_y.
\label{eq:project}
\end{equation}
Visibility is determined by comparing the projected depth against the rendered depth buffer $D_t$:
\begin{equation}
\mathrm{visible}_{t} = \mathbf{1}\!\left[\, p^{\mathrm{cam}}_{t,z} - D_t(u_{t}, v_{t}) < \epsilon \,\right],
\label{eq:visibility}
\end{equation}
where $\epsilon = 0.03\,\text{m}$.

\begin{table}[ht!]
\centering
\caption{\textbf{Statistics of \benchmarkdata{}.} Every episode is recorded by an exo and a wrist camera. Static tracks move less than one pixel (at 256-pixel image height) from their first visible position over the whole clip. A point counts as not visible when it is occluded or outside the frame.}
\label{tab:data_stats}
\small
\setlength{\tabcolsep}{5pt}
\begin{tabular}{l rrr}
\toprule
Houses / episodes           & \multicolumn{3}{r}{300 / 1,500} \\
Videos                      & \multicolumn{3}{r}{3,000 (1,500 per camera)} \\
Tracks                      & \multicolumn{3}{r}{15M (5,000 per video)} \\
Frames / video              & \multicolumn{3}{r}{70 on average (42--177)} \\
Resolution                  & \multicolumn{3}{r}{$624\times352$, 15\,fps} \\
\midrule
 & \textbf{Exo} & \textbf{Wrist} & \textbf{All} \\
\cmidrule(lr){2-4}
Tracks on the robot         & 82.1\% & 69.6\% & 75.8\% \\
Tracks on objects           & 17.9\% & 30.4\% & 24.2\% \\
Queried on the first frame  & 61.8\% & 77.6\% & 69.7\% \\
Static tracks               & 16.7\% & 12.6\% & 14.7\% \\
Not visible at least once   & 13.5\% & 16.5\% & 15.0\% \\
\bottomrule
\end{tabular}
\end{table}

\section{Training CoWTracker}
\label{appendix:cowtracker_control}

\modelname{}-Warp is CoWTracker~\citep{lai2026cowtracker} fine-tuned for 10k steps at learning rate $10^{-5}$ on an equal mix of Kubric and \benchmarkdata{}. CoWTracker's training code wasn't released, so we trained using our own re-implementation. To separate the effect of \benchmarkdata{} from the effect of the additional training, we fine-tune the released CoWTracker with the same recipe on Kubric alone (\Cref{tab:cowtracker_control}). Compared with this control, \benchmarkdata{} adds 1.2 AJ on \benchmarkreal{} and \proto{0.4}{0.5} AJ on RoboTAP and on the robotics average, with small gains on Kinetics (+0.1) and DAVIS (+0.3) and a small loss on RGB-Stacking ($-$0.3). The remaining 0.9 AJ of the 2.1 AJ gain over the released model on \benchmarkreal{} comes from the additional Kubric training. The lower occlusion accuracy on DAVIS relative to the released model also appears in the control, so it stems from the fine-tuning recipe rather than from \benchmarkdata{}.

\ifallpoints\begin{table*}[t]
\centering
\caption{\textbf{Kubric-only control for \modelname{}-Warp.} All three are models using the CoWTracker architecture. Avg.\ is the mean over the three robotics benchmarks. Evaluation as in \cref{tab:results}. Bold marks the best value in each column.}
\label{tab:cowtracker_control}
\resizebox{\textwidth}{!}{
\begin{tabular}{l ccc ccc ccc ccc @{\hspace{1.5em}} ccc ccc}
\toprule
& \multicolumn{3}{c}{\textbf{\benchmarkreal{}}} & \multicolumn{3}{c}{\textbf{RGB-Stack}} & \multicolumn{3}{c}{\textbf{RoboTAP}} & \multicolumn{3}{c}{\textbf{Avg.}} & \multicolumn{3}{c}{\textbf{Kinetics}} & \multicolumn{3}{c}{\textbf{DAVIS}} \\
\cmidrule(lr){2-4} \cmidrule(lr){5-7} \cmidrule(lr){8-10} \cmidrule(lr){11-13} \cmidrule(lr){14-16} \cmidrule(lr){17-19}
Training data & AJ & $\delta_{\text{vis}}$ & OA & AJ & $\delta_{\text{vis}}$ & OA & AJ & $\delta_{\text{vis}}$ & OA & AJ & $\delta_{\text{vis}}$ & OA & AJ & $\delta_{\text{vis}}$ & OA & AJ & $\delta_{\text{vis}}$ & OA \\
\midrule
Released CoWTracker (Kubric)
& 63.9 & 76.1 & 88.5
& 85.5 & 93.0 & \textbf{94.8}
& 70.5 & 81.2 & 93.9
& 73.3 & 83.4 & 92.4
& 57.6 & 69.9 & \textbf{90.0}
& \textbf{66.1} & 78.4 & \textbf{91.2} \\
\quad + Kubric only (control)
& 64.8 & 77.1 & 88.1
& \textbf{86.2} & 93.0 & 94.6
& 71.1 & 81.6 & 93.9
& 74.0 & 83.9 & 92.2
& 57.9 & 70.2 & 89.4
& 65.6 & 78.6 & 89.8 \\
\quad + Kubric + \benchmarkdata{} (\modelname{}-Warp)
& \textbf{66.0} & \textbf{77.9} & \textbf{89.0}
& 85.9 & \textbf{93.2} & 94.5
& \textbf{71.6} & \textbf{81.9} & \textbf{94.2}
& \textbf{74.5} & \textbf{84.4} & \textbf{92.6}
& \textbf{58.0} & \textbf{70.3} & 89.7
& 65.9 & \textbf{78.9} & 89.9 \\
\bottomrule
\end{tabular}}
\end{table*}

\else\begin{table*}[t]
\centering
\caption{\textbf{Kubric-only control for \modelname{}-Warp.} All three are models using the CoWTracker architecture. Avg.\ is the mean over the three robotics benchmarks. Evaluation as in \cref{tab:results}. Bold marks the best value in each column.}
\label{tab:cowtracker_control}
\resizebox{\textwidth}{!}{
\begin{tabular}{l ccc ccc ccc ccc @{\hspace{1.5em}} ccc ccc}
\toprule
& \multicolumn{3}{c}{\textbf{\benchmarkreal{}}} & \multicolumn{3}{c}{\textbf{RGB-Stack}} & \multicolumn{3}{c}{\textbf{RoboTAP}} & \multicolumn{3}{c}{\textbf{Avg.}} & \multicolumn{3}{c}{\textbf{Kinetics}} & \multicolumn{3}{c}{\textbf{DAVIS}} \\
\cmidrule(lr){2-4} \cmidrule(lr){5-7} \cmidrule(lr){8-10} \cmidrule(lr){11-13} \cmidrule(lr){14-16} \cmidrule(lr){17-19}
Training data & AJ & $\delta_{\text{vis}}$ & OA & AJ & $\delta_{\text{vis}}$ & OA & AJ & $\delta_{\text{vis}}$ & OA & AJ & $\delta_{\text{vis}}$ & OA & AJ & $\delta_{\text{vis}}$ & OA & AJ & $\delta_{\text{vis}}$ & OA \\
\midrule
Released CoWTracker (Kubric)
& 63.7 & 75.9 & 88.6
& 85.5 & 93.0 & \textbf{94.8}
& 73.5 & 83.7 & 94.6
& 74.2 & 84.2 & 92.7
& 60.8 & 73.1 & \textbf{91.5}
& \textbf{66.1} & 78.4 & \textbf{91.2} \\
\quad + Kubric only (control)
& 64.5 & 76.9 & 88.1
& \textbf{86.2} & 93.0 & 94.6
& 73.9 & 83.8 & 94.7
& 74.9 & 84.5 & 92.5
& 60.9 & 73.2 & 91.0
& 65.6 & 78.6 & 89.8 \\
\quad + Kubric + \benchmarkdata{} (\modelname{}-Warp)
& \textbf{65.8} & \textbf{77.7} & \textbf{89.1}
& 85.9 & \textbf{93.2} & 94.5
& \textbf{74.3} & \textbf{84.3} & \textbf{94.8}
& \textbf{75.3} & \textbf{85.1} & \textbf{92.8}
& \textbf{61.1} & \textbf{73.4} & 91.3
& 65.9 & \textbf{78.9} & 89.9 \\
\bottomrule
\end{tabular}}
\end{table*}
\fi

\section{Results by Video Source}
\label{appendix:results_source}

\begin{figure}[t]
    \centering
    \includegraphics[width=0.7\linewidth]{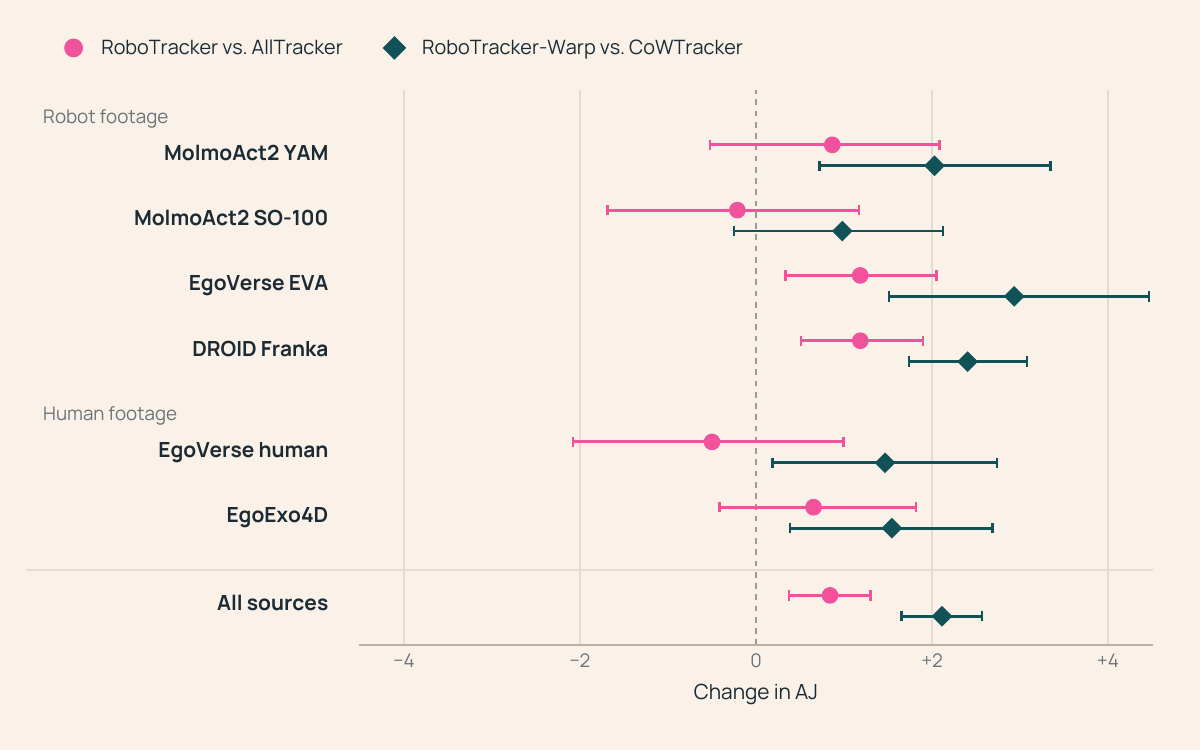}
    \caption{\textbf{Gains on \benchmarkreal{} by video source.} Mean per-clip change in AJ of each of our models over the model it was fine-tuned from, with 95\% confidence intervals (paired bootstrap over clips). Per-source scores for all models are in \Cref{tab:per_source}.}
    \label{fig:per_source_gains}
\end{figure}
\begin{table*}[t]
\centering
\caption{\textbf{Results on \benchmarkreal{} by video source.} Each source's score is the mean over its clips, the same aggregation as the overall score (All); clip counts in parentheses. Evaluation as in \Cref{tab:results}. Bold marks the best and underline the second-best value in each column; our models are shaded.}
\label{tab:per_source}
\setlength{\tabcolsep}{3.5pt}
\resizebox{\textwidth}{!}{
\begin{tabular}{l >{\footnotesize}l ccc ccc ccc ccc ccc ccc ccc}
\toprule
& & \multicolumn{3}{c}{\textbf{DROID}} & \multicolumn{6}{c}{\textbf{MolmoAct2}} & \multicolumn{6}{c}{\textbf{EgoVerse}} & \multicolumn{3}{c}{\textbf{Ego-Exo4D}} & \multicolumn{3}{c}{\textbf{All}} \\
\cmidrule(lr){3-5} \cmidrule(lr){6-11} \cmidrule(lr){12-17} \cmidrule(lr){18-20} \cmidrule(lr){21-23} 
& & \multicolumn{3}{c}{Franka (167)} & \multicolumn{3}{c}{YAM (24)} & \multicolumn{3}{c}{SO-100 (24)} & \multicolumn{3}{c}{EVA (25)} & \multicolumn{3}{c}{Human (25)} & \multicolumn{3}{c}{Human (42)} & \multicolumn{3}{c}{All sources (307)} \\
\cmidrule(lr){3-5} \cmidrule(lr){6-8} \cmidrule(lr){9-11} \cmidrule(lr){12-14} \cmidrule(lr){15-17} \cmidrule(lr){18-20} \cmidrule(lr){21-23}
Model & \multicolumn{1}{l}{Train. Data} & AJ & $\delta_{\text{vis}}$ & OA & AJ & $\delta_{\text{vis}}$ & OA & AJ & $\delta_{\text{vis}}$ & OA & AJ & $\delta_{\text{vis}}$ & OA & AJ & $\delta_{\text{vis}}$ & OA & AJ & $\delta_{\text{vis}}$ & OA & AJ & $\delta_{\text{vis}}$ & OA \\
\midrule
CoTracker3 (offline) & Kubric
& 60.0 & 73.6 & 84.4
& 68.3 & 80.0 & 89.6
& 58.1 & 73.2 & 85.6
& 61.8 & 77.2 & 86.0
& 58.3 & 72.4 & 82.4
& 54.7 & 66.1 & 82.5
& 59.8 & 73.2 & 84.6 \\
CoTracker3 (offline) & Kubric+15k
& 60.1 & 72.6 & 86.0
& 66.8 & 77.8 & 88.5
& 57.8 & 72.3 & 86.1
& 63.1 & 75.6 & 88.6
& 61.0 & 74.0 & 85.3
& 56.2 & 67.0 & 82.2
& 60.2 & 72.6 & 85.8 \\
CoTracker3 (online) & Kubric+15k
& 61.8 & 74.8 & 86.9
& 68.3 & \underline{81.0} & 90.3
& 59.2 & 74.8 & 87.4
& 64.5 & 79.1 & 87.6
& 62.7 & \textbf{75.6} & \textbf{89.2}
& 57.8 & 68.9 & \underline{85.9}
& 61.8 & 74.9 & 87.3 \\
TAPNext & Kubric
& 56.6 & 71.8 & 87.7
& 59.1 & 73.2 & 90.6
& 53.9 & 69.2 & 88.6
& 55.1 & 71.0 & 87.5
& 48.0 & 62.1 & 82.3
& 51.3 & 63.7 & 84.9
& 55.1 & 69.7 & 87.2 \\
BootsTAPNext & Kubric+15M
& 62.3 & 75.9 & \underline{88.7}
& 64.0 & 76.6 & 89.1
& 59.0 & 72.5 & \underline{89.0}
& 59.4 & 74.3 & 86.5
& 50.9 & 64.3 & 80.4
& 56.0 & 67.6 & 83.3
& 60.2 & 73.5 & 87.1 \\
AllTracker & Kubric
& 60.2 & 73.9 & 82.4
& 68.7 & 79.9 & 89.0
& 61.1 & 74.2 & 86.0
& 66.7 & 79.5 & 88.2
& \underline{62.8} & 75.3 & 87.1
& 58.1 & 70.5 & 81.2
& 61.4 & 74.5 & 83.9 \\
AllTracker & Kubric+mix
& 62.0 & 75.6 & 86.0
& 68.9 & 80.4 & 91.1
& 61.4 & \underline{76.0} & 87.2
& 66.2 & 79.2 & 89.3
& \textbf{62.9} & \textbf{75.6} & \underline{88.8}
& 59.6 & 71.1 & 84.9
& 62.6 & 75.7 & 86.9 \\
CoWTracker & Kubric
& \underline{64.4} & \underline{77.0} & 88.2
& 69.2 & 80.2 & \underline{92.1}
& \underline{63.2} & 75.2 & \textbf{89.4}
& 66.4 & 79.2 & \textbf{90.8}
& 60.9 & 73.7 & 87.7
& 59.2 & 70.6 & \textbf{86.0}
& \underline{63.9} & \underline{76.1} & \underline{88.5} \\
\midrule
\rowcolor{oursbg}\modelname{} & Kubric+mix+\benchmarkdata{}
& 63.2 & 76.3 & 85.8
& \underline{69.7} & 80.4 & 91.1
& 61.2 & 75.3 & 85.8
& \underline{67.4} & \underline{79.9} & 89.0
& 62.4 & \underline{75.4} & 86.8
& \underline{60.3} & \underline{71.8} & 83.2
& 63.4 & \underline{76.1} & 86.2 \\
\rowcolor{oursbg}\modelname{}-Warp & Kubric+\benchmarkdata{}
& \textbf{66.8} & \textbf{78.8} & \textbf{89.7}
& \textbf{71.3} & \textbf{81.9} & \textbf{92.3}
& \textbf{64.2} & \textbf{76.7} & \underline{89.0}
& \textbf{69.4} & \textbf{82.3} & \underline{90.3}
& 62.4 & 74.9 & 87.1
& \textbf{60.8} & \textbf{72.1} & 84.8
& \textbf{66.0} & \textbf{77.9} & \textbf{89.0} \\
\bottomrule
\end{tabular}}
\end{table*}

\Cref{tab:per_source} breaks down \benchmarkreal{} results by video source, splitting MolmoAct2 into its YAM and SO-100 clips and EgoVerse into its robot and human clips. Human manipulation is the hardest setting: every model scores its lowest AJ on either Ego-Exo4D or the human EgoVerse clips. The gains from \benchmarkdata{}, which contains only robot manipulation, are concentrated on robot videos. \modelname{}-Warp improves over CoWTracker on every source (1.0 to 3.0 AJ) and is the best model on all sources except the human EgoVerse clips, where AllTracker (Kubric+mix) leads. \modelname{} improves over its starting model, AllTracker (Kubric+mix), on DROID, the YAM clips, and the EgoVerse robot clips (+0.8 to +1.2 AJ) and on Ego-Exo4D (+0.7), is comparable on the SO-100 clips, and is 0.5 AJ lower on the human EgoVerse clips. Rankings among off-the-shelf models also change between sources: BootsTAPNext slightly outperforms CoTracker3 (online) on DROID but falls 11.9 AJ behind AllTracker (Kubric) on the human EgoVerse clips. \Cref{fig:per_source_gains} shows the same gains as mean per-clip changes with 95\% confidence intervals. Because every subset other than DROID contains only 24 to 42 clips, small per-source differences should be interpreted with caution.

\section{Qualitative Examples}
\label{appendix:qual_examples}

\subsection{\modelname{}-Warp Predictions}

\begin{figure*}[t]
  \centering
  \setlength{\tabcolsep}{0pt}
  \begin{tabular}{@{}c@{\hspace{0.03\textwidth}}c@{}}
    \includegraphics[width=0.47\textwidth]{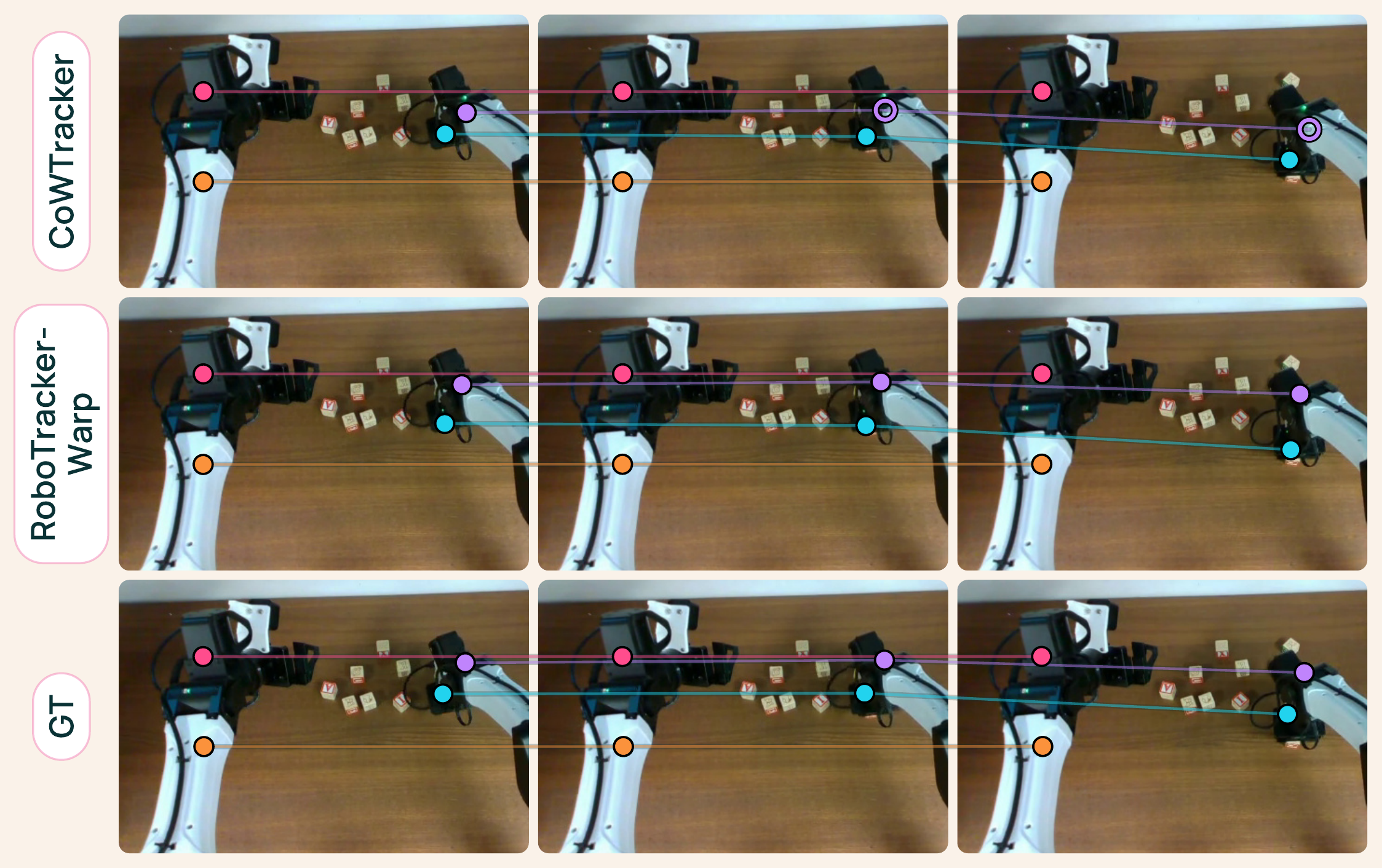} &
    \includegraphics[width=0.47\textwidth]{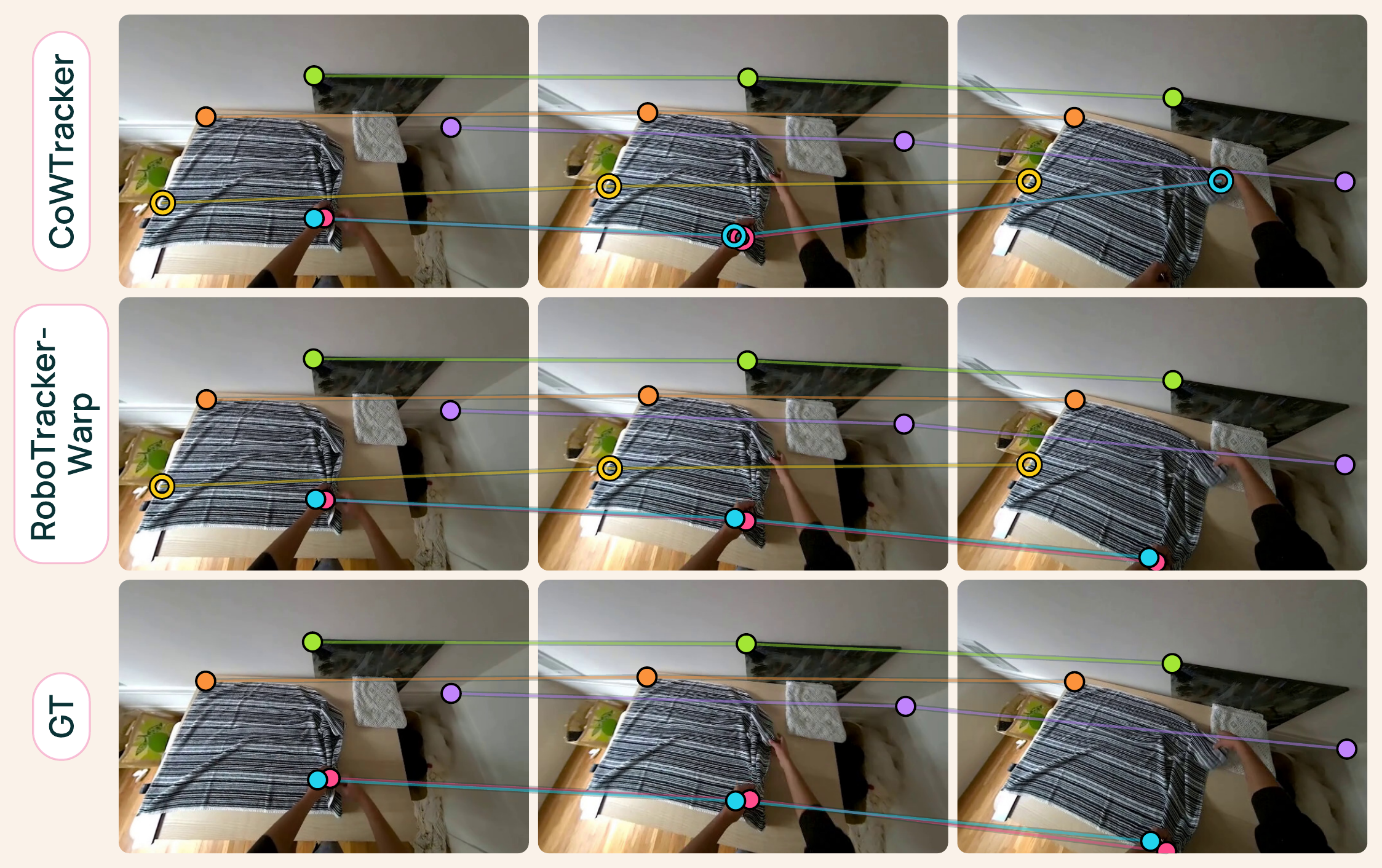} \\
    \small (a) MolmoAct2 & \small (b) EgoVerse \\[6pt]
    \includegraphics[width=0.47\textwidth]{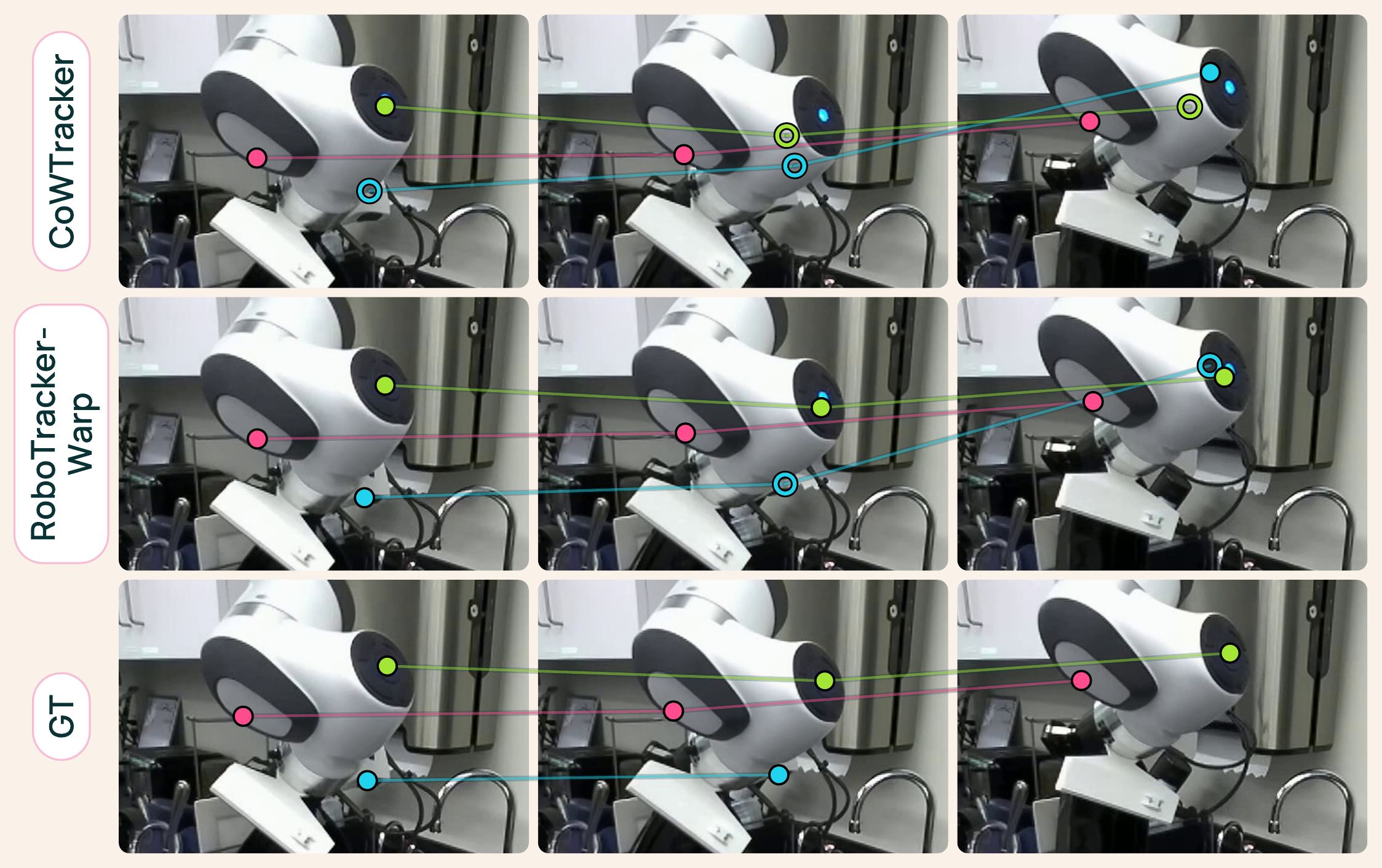} &
    \includegraphics[width=0.47\textwidth]{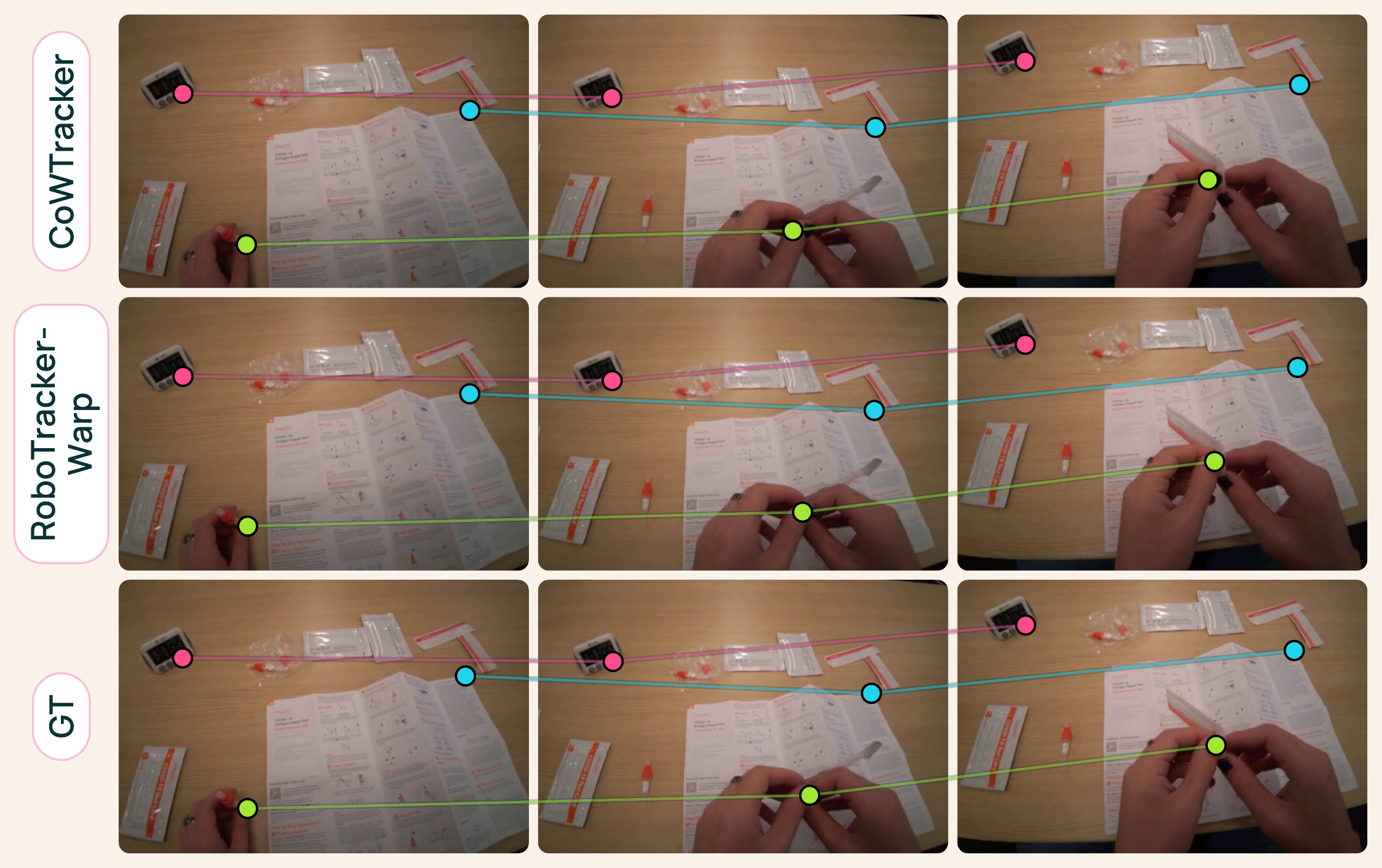} \\
    \small (c) DROID & \small (d) Ego-Exo4D
  \end{tabular}
  \caption{\textbf{CoWTracker vs.\ \modelname{}-Warp on \benchmarkreal{}.} Each panel shows CoWTracker, \modelname{}-Warp, and the ground truth (GT) in rows, and three frames in temporal order in columns. Filled dots are visible points and hollow points are predicted to be occluded; each color is one track. CoWTracker loses points as the robot, hands, and objects move, while \modelname{}-Warp stays on them.}
  \label{fig:warp_qual}
\end{figure*}

\Cref{fig:warp_qual} compares \modelname{}-Warp with the released CoWTracker on \benchmarkreal{} clips where fine-tuning on \benchmarkdata{} gives a large gain, one from each source.

\subsection{Failure Cases}
\label{appendix:failure_cases}

\begin{figure*}[t]
  \centering
  \setlength{\tabcolsep}{0pt}
  \begin{tabular}{@{}c@{\hspace{0.03\textwidth}}c@{}}
    \includegraphics[width=0.47\textwidth]{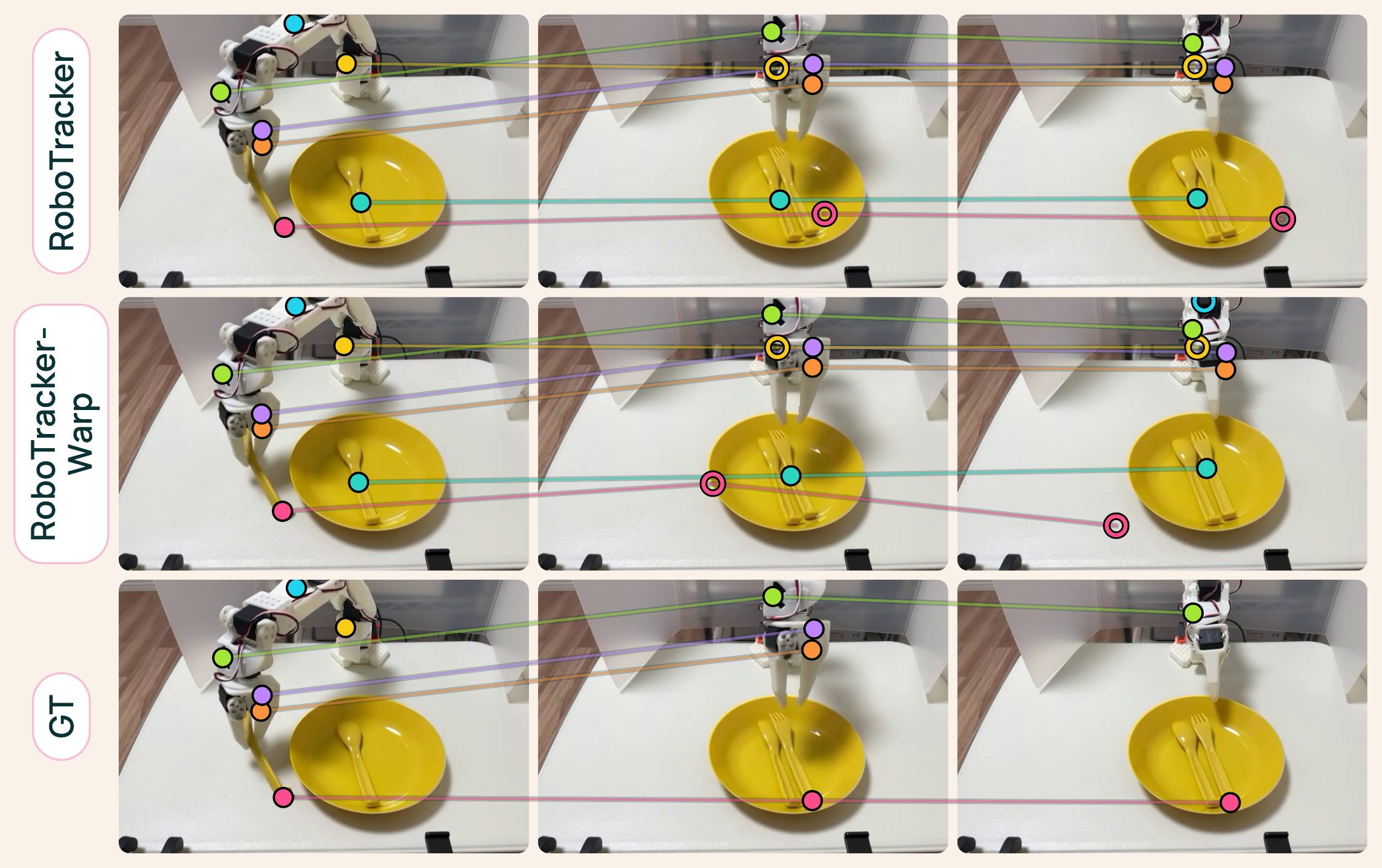} &
    \includegraphics[width=0.47\textwidth]{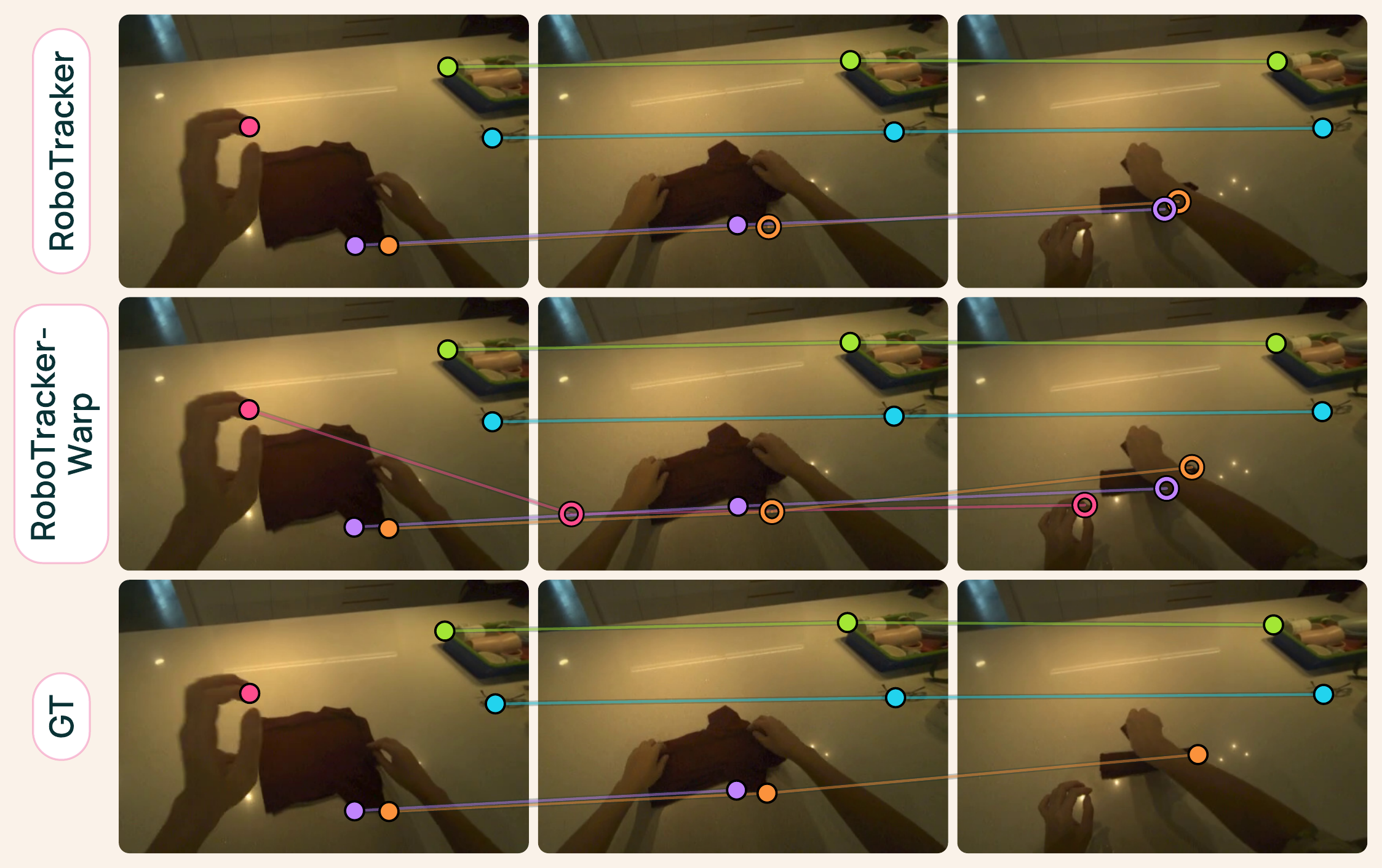} \\
    \small (a) MolmoAct2 & \small (b) EgoVerse \\[6pt]
    \includegraphics[width=0.47\textwidth]{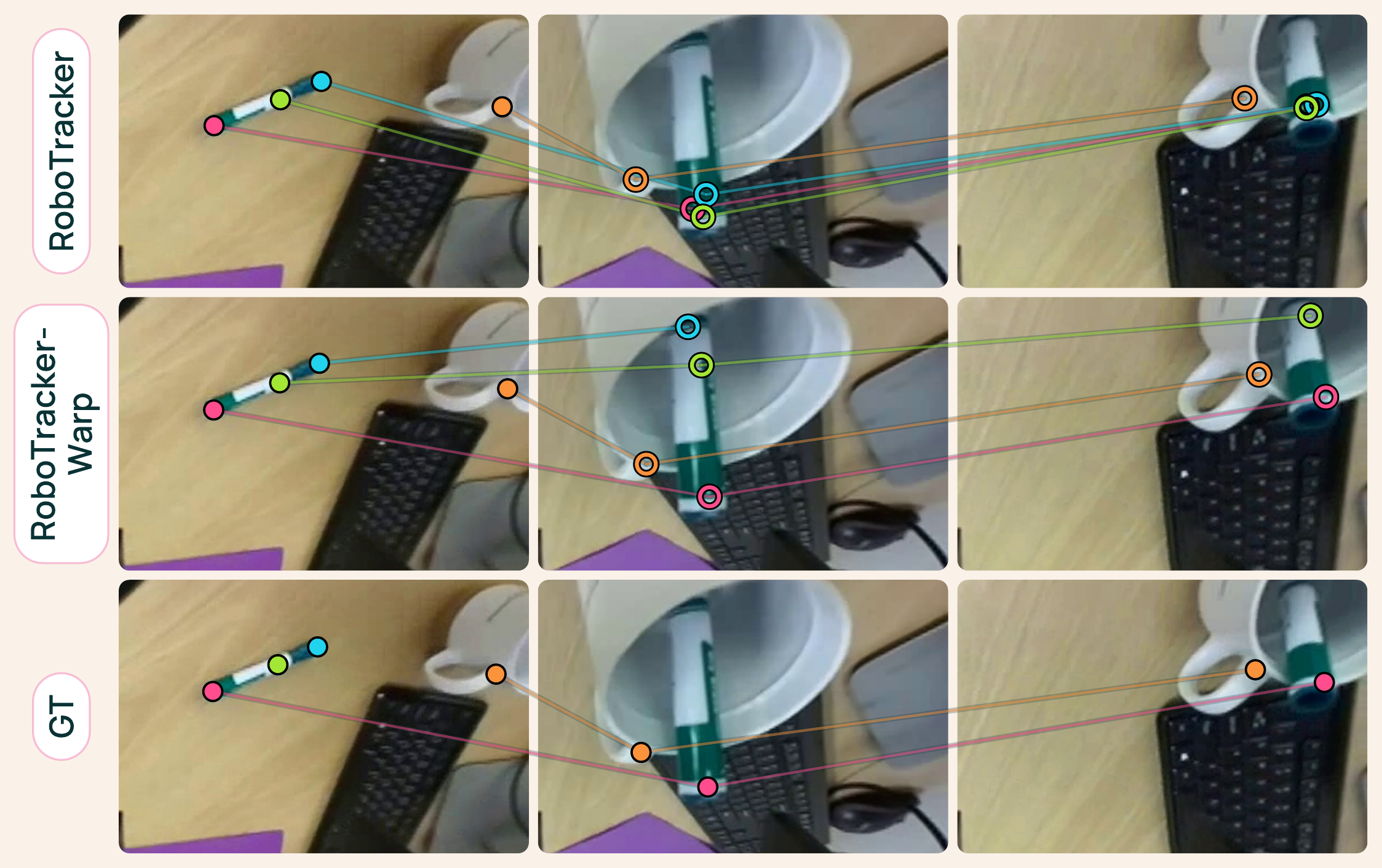} &
    \includegraphics[width=0.47\textwidth]{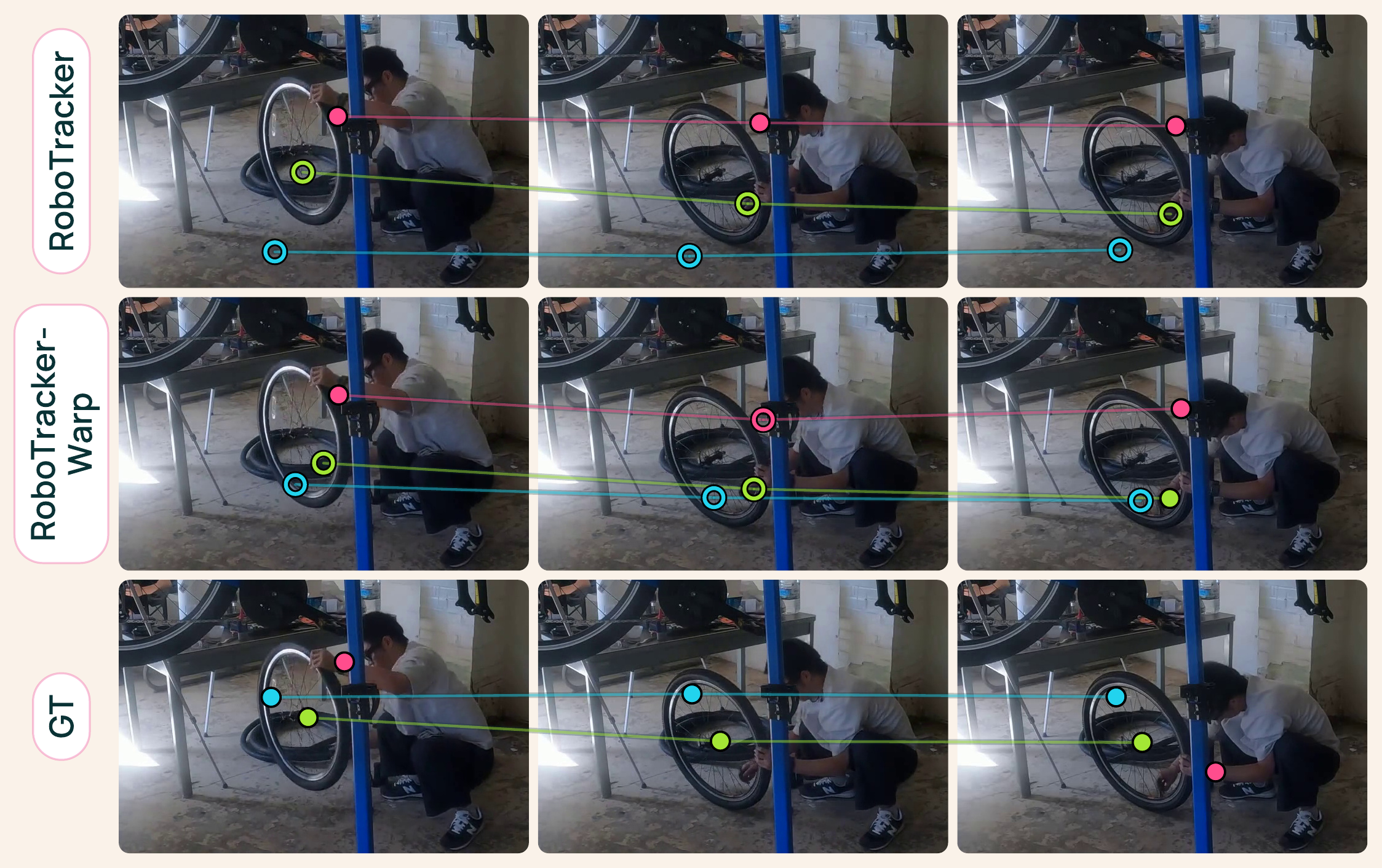} \\
    \small (c) DROID & \small (d) Ego-Exo4D
  \end{tabular}
  \caption{\textbf{Failure cases shared by \modelname{} and \modelname{}-Warp.} Each panel shows \modelname{}, \modelname{}-Warp, and the ground truth (GT) in rows, and three frames in temporal order in columns: before, when, and after both models lose a point. Filled dots are visible points and hollow points are predicted to be occluded; each color is one track. Both models lose points on small or fast-moving objects and on hands, such as the cutlery lifted by the SO-100 arm (a), a cloth handled by both hands (b), a marker placed into a cup (c), and a turning bicycle wheel (d).}
  \label{fig:failure_qual}
\end{figure*}

\Cref{fig:failure_qual} shows clips where both \modelname{} and \modelname{}-Warp fail, one from each source. In each, a point both models track correctly is lost once the object moves quickly, turns, or is covered by a hand or gripper.

\section{Limitations}
\label{appendix:limitation}

\benchmarkreal{} contains 307 clips with 2,522 annotated tracks, which is modest in scale. Acquiring more annotations is costly because each track requires careful human verification, and ensuring trajectory correctness through occlusions remains inherently ambiguous even for annotators. Although the clips span several sources and embodiments, they do not cover all manipulation skills; for example, tasks involving deformable objects or tool use are underrepresented. We fine-tune two trackers, AllTracker and CoWTracker. Although these are the two strongest architecture families, training on more model types would let us be more confident in \benchmarkdata{}'s benefit. Additionally, CoWTracker did not release its training code, so we re-implemented our version and cannot say for certain that behavior is the exact same. Other recent trackers do not release training pipelines or use bootstrapped training procedures that are difficult to replicate, so we cannot measure how much \benchmarkdata{} would benefit those models. 

Despite improvements from \benchmarkdata{}, important challenges remain. \benchmarkdata{} improves sim-to-real transfer but does not close the gap entirely, suggesting that scene diversity, rendering realism, or motion distribution in our synthetic data still diverge from real manipulation in ways we have not fully characterized. Our qualitative results show that models still struggle with long occlusions, reappearance after occlusion, and small objects that can be confused with the background. These limitations suggest promising directions for future work, including improved occlusion reasoning, higher-resolution tracking for small objects, and richer simulation data with more complex object interactions. We hope that \benchmarkreal{} will support future progress toward reliable point tracking for real-world robotic manipulation.

\section{Broader Impact}
\label{appendix:broader_impact}

Point tracking is a general-purpose motion estimation primitive with applications well beyond robotic manipulation. On the positive side, more reliable point trackers can improve assistive and household robotics, where accurate motion estimation is needed for safe interaction with people and objects. The same capability is useful in ecological research, for instance tracking animal movement in wildlife footage, and in medical settings such as surgical video analysis, where instrument and tissue motion must be followed precisely.

On the negative side, improvements in point tracking could be repurposed for surveillance, enabling more robust tracking of individuals across camera views or in crowded environments without their consent. More capable tracking models could also be integrated into autonomous weapons systems, where precise motion estimation of targets is operationally relevant. We note that our contribution is a benchmark and training data rather than a deployed tracking system, which limits its direct potential for misuse, but acknowledge that downstream improvements enabled by this work could be applied in such contexts.

\section{Licenses}
\label{appendix:licenses}

DROID~\citep{khazatsky2024droid} is released under CC-BY 4.0. Ego-Exo4D~\citep{grauman2024egoexo4d} is released under the Ego-Exo4D license agreement. EgoVerse is released under CC-BY-SA 4.0. AllTracker is released under MIT. Kubric~\citep{greff2022kubric} is released under Apache 2.0. MolmoSpaces~\citep{kim2026molmospaces} data is released under CC-BY 4.0 (with Objaverse subsets under ODC-BY 1.0) and code under Apache 2.0. CoTracker3~\citep{cotracker3} is released under CC-BY-NC 4.0. CoWTracker weights are released under CC-BY-NC 4.0. TAP-Vid~\citep{tapvid} annotations, RGB-Stacking videos, and RoboTAP~\citep{vecerik2024robotap} videos are released under CC-BY. The PointOdyssey~\citep{zheng2023pointodyssey} dataset is released under CC-BY-NC-SA 4.0. We respect all license terms in our use of these assets.

\end{document}